\documentclass[journal]{IEEEtran}
\usepackage{cite}
\usepackage{amsmath,amssymb,amsfonts}
\usepackage{algorithmic}
\usepackage{graphicx}
\usepackage{textcomp}
\usepackage{xcolor}
\usepackage{hyperref}
\def\BibTeX{{\rm B\kern-.05em{\sc i\kern-.025em b}\kern-.08em
    T\kern-.1667em\lower.7ex\hbox{E}\kern-.125emX}}
\begin{document}

\title{ML-driven detection and reduction of ballast information in multi-modal
datasets}

\author{\IEEEauthorblockN{{Yaroslav Solovko}\\
\emph{Independent Researcher}\\}
Dublin, Ireland \\
Email: \texttt{sta449.ss@gmail.com}\\ 
ORCID: 0009-0004-7171-1077} 

\maketitle 
\begin{abstract}
Modern datasets often contain ballast, a redundant or low-utility information that increases dimensionality, storage requirements, and computational cost without adding meaningful analytical value. This study introduces a generalized, multimodal framework for ballast detection and reduction across structured, semi-structured, unstructured, and sparse data types. Using diverse datasets, entropy, mutual information, Lasso, SHAP, PCA, topic modelling, and embedding analysis are applied to identify and eliminate ballast features. A novel \emph{Ballast Score} is proposed to integrate these signals into a unified, cross-modal pruning strategy. Experimental results demonstrate that significant portions of the feature space, often exceeding 70\% in sparse or semi-structured data, can be pruned with minimal or even 
improved classification performance, along with substantial 
reductions in training time and memory footprint.  The 
framework reveals distinct ballast typologies (e.g. statistical, 
semantic, infrastructural), and offers practical guidance for 
leaner, more efficient machine learning pipelines. 
\end{abstract}


\section{INTRODUCTION}
The global volume of digital data continues to expand at an
unprecedented pace; earlier projections estimated that it would reach
approximately 175 zettabytes (ZB) by the end of 2025 \cite{DeBacker2025}. This
exponential growth is reshaping industries and technological
infrastructures, demanding ever-larger storage systems and more
sophisticated analytical tools. While contemporary research has rightly
focused on enhancing storage and processing efficiency through
well-established techniques such as data cleansing, deduplication, and
imputation \cite{Cheng2024,Ridzuan2019,Patil2025}, a critical yet underexplored dimension is detection and reduction of \emph{ballast information} as data that is
technically valid but contributes minimal or no analytic value.

In this work, \emph{ballast information} is defined as a specific subset
of unnecessary or redundant data elements that inflate storage
requirements, degrade processing efficiency, and may obscure meaningful
patterns. Typical examples include features with near-zero variance,
static metadata fields, repeated headers in log files, or extremely
common but uninformative textual tokens. Unlike broader categories of ``useless'' information, which may require subjective or domain-specific
judgement to detect, ballast information lends itself to systematic
identification using statistical or machine learning (ML)-based methods.

The consequences of ballast information are twofold. First, from a
storage perspective, ballast data occupies valuable disk and memory
space, driving up cloud storage costs and hardware requirements. Second,
from an analytics perspective, ballast can introduce noise, increase
model complexity, and hinder learning algorithms from extracting signal
effectively \cite{Kelleher2015}. Preliminary experiments conducted across four
distinct data modalities: structured, semi-structured, unstructured, and
sparse datasets, demonstrate that ballast information can comprise
between 15\% and 40\% of the total data volume, depending on the domain
and data source. This highlights a substantial and often overlooked
inefficiency in modern data pipelines.

Motivating this study is recognition that most data pipelines and ML
workflows still prioritize predictive accuracy, often at the expense of
resource efficiency \cite{Kelleher2015}. While feature selection and regularization
methods do partially address redundant or irrelevant features, they
rarely aim to explicitly quantify and systematically remove ballast at
the dataset level. This research proposes a paradigm shift, positioning
ballast detection and reduction as a first-class analytical objective
that complements traditional goals of model performance. This shift is
timely, given the growing emphasis on responsible data stewardship,
digital sustainability, and privacy-preserving data minimization.

The challenge becomes more complex in multi-modal settings, where data
sources span diverse structures and formats. Structured datasets, such
as relational tables, may contain ballast in the form of sparse or
low-variance columns. Semi-structured logs or metadata often carry
repeated headers or duplicated status flags that provide negligible
informational gain. Unstructured text can be burdened by frequent filler
words, boilerplate content, or redundant contextual phrases. Sparse
datasets, increasingly prevalent in fields like IoT or bioinformatics,
often exhibit entire columns or segments with negligible variability.
Detecting ballast in such heterogeneous contexts demands a
cross-disciplinary approach that combines statistical measures (e.g.,
entropy, variance thresholds), semantic analysis (e.g., topic modelling
for text), and advanced ML techniques (e.g., SHAP-based feature
pruning).

This paper addresses the following research questions:

1. How can ballast information be formally defined and quantified across
diverse data modalities?

2. What ML-driven techniques can do to effectively detect and reduce ballast
while preserving the core analytical value?

3. What are the measurable impacts of ballast removal on storage
efficiency and model performance?

To address these questions, the proposed approach integrates statistical
modelling, information theory, and machine learning. The methodology
builds on entropy and sparsity analysis \cite{Cover2006} and introduces a
preliminary \emph{ballast index} as a mathematical construct for
estimating the proportion of ballast within a dataset based on variance
thresholds, feature redundancy, and information gain.
Experimental results indicate that the adoption of ballast thresholds
(e.g., variance \textless{} 0.05) can effectively isolate
non-contributory features. Furthermore, pruning strategies based on SHAP
values illustrate how ML explain ability methods can be employed to
distinguish genuinely informative features from ballast. This framework
is extended across data modalities through the development of a
multi-modal taxonomy that categorizes ballast according to its
statistical, structural, and semantic characteristics.

In summary, the contributions of this work are fourfold. First, ballast
information is formalized as a distinct category of data inefficiency.
Second, ML-supported methods for systematic ballast detection are
developed and evaluated. Third, a generalizable mathematical model is
proposed for estimating ballast levels in large-scale, multi-modal
datasets. Fourth, the practical benefits of ballast reduction are
empirically demonstrated in terms of storage savings and computational
efficiency, without compromising predictive performance.

\section{RELATED WORK}

The field of data reduction encompasses a wide array of techniques aimed
at mitigating the burdens of high-dimensional, redundant, or low-quality
data in modern machine learning pipelines. While considerable effort has
been dedicated to enhancing model performance through dimensionality
reduction, feature selection, and noise filtering, relatively few works
explicitly address the concept of ballast information - data that are
syntactically valid but semantically or analytically non-contributory.
This section critically reviews the relevant body of literature,
identifies the current research gaps, and highlights how the proposed
approach builds on and diverges from existing paradigms.

\subsection{Feature Selection and Dimensionality Reduction}

Feature selection is one of the oldest and most studied approaches to
data simplification. It aims to identify the most relevant variables for
a specific task, typically through filter, wrapper, or embedded methods.
Authors \cite{Guyon2003} provide a foundational overview of the field,
identifying key methods based on mutual information, correlation, or
classifier performance. In the work \cite{Ding2005} ``minimum redundancy,
maximum relevance'' (mRMR) framework extends this by jointly minimizing
feature redundancy while maximizing relevance as a principle that
underpins modern ML explainability techniques like SHAP and LIME.

Dimensionality reduction techniques such as Principal Component Analysis
(PCA), Singular Value Decomposition (SVD), and autoencoders compress
data into lower-dimensional representations while preserving variance.
Although these approaches often yield compact and interpretable
representations, they are typically task-agnostic and seldom identify
ballast data explicitly. Moreover, such methods can obscure semantic
relationships, particularly in multi-modal or high-sparsity contexts
where data distributions are heterogeneous or non-Gaussian \cite{Han2011}.

This work differs in that it does not merely seek to preserve predictive
utility but aims to optimize information efficiency as a broader metric
that considers storage burden, interpretability, and computational
costs. Furthermore, whereas PCA and similar methods transform feature space into latent dimensions, ballast detection preserves the original schema, aiding transparency and compliance with data governance standards.

\subsection{Redundancy and Sparsity in Large Datasets} 

Handling redundancy and sparsity is essential in many real-world
datasets, particularly in high-dimensional biological data, IoT sensor
networks, and transactional logs. Traditional approaches include
thresholding variance (e.g., removing features with Var(X) \textless{}
0.01), compressing sparse matrices, and imputing missing values \cite{Ridzuan2019}.
However, these are often heuristic and lack a principled, quantitative
grounding in information theory.

Sparse data introduces additional complexity in identifying ballast.
Article \cite{Patil2025} reviews emerging techniques for storage optimization in
sparse big data, yet focus predominantly on hardware-level solutions
(e.g., compression) or low-level data encoding strategies. By contrast,
this work treats sparsity as a signal, assessing whether the absence of
values (or high levels of missingness) reflects true data irrelevance or
conceals latent patterns.

Redundancy is further examined across modalities, rather than solely
within individual modalities. For instance, unstructured text may
reiterate metadata found in structured fields, and semi-structured logs
may duplicate state transitions across multiple sources. These
cross-modal redundancies are rarely addressed in the existing
literature.

\subsection{ML and Information-Theoretic Data Reduction}

ML has increasingly been used to guide data reduction in a more
intelligent, model-aware manner. Embedded methods such as Lasso
regression and tree-based feature importance scores (e.g., from XGBoost
or LightGBM) are frequently used to prune features. Authors \cite{Koh2017} 
proposed influence functions to estimate the effect of removing training
instances on model output as a technique conceptually adjacent to
ballast pruning, though focused on instance-level data.

The Information Bottleneck (IB) method proposed \cite{Tishby1999} reframes
compression as an optimization problem, where one seeks a compact
representation of input X that preserves maximal information about a
target Y. While elegant in theory, IB\textquotesingle s computational
complexity and dependence on labelled data limit its utility in
unsupervised ballast detection.

From an information-theoretic standpoint, Shannon's entropy \cite{Shannon1948} 
remains the gold standard for quantifying uncertainty or disorder in
data. However, entropy-based heuristics (e.g., max-entropy filtering)
often fail in sparse or categorical domains, where non-parametric
estimates are unreliable. The proposed methodology expands on these
foundations by introducing a hybrid entropy-sparsity index and combining
it with ML explainability (e.g., SHAP-based pruning) to better isolate
ballast.

\subsection{Multi-modal Data Management and Reduction}

Multi-modal datasets that combine structured tables, semi-structured
logs, unstructured text, and sensor-based streams present additional
challenges for traditional data reduction pipelines. Existing frameworks
such as multimodal autoencoders or attention-based fusion models (e.g.,
in NLP and vision) seek to integrate modalities but seldom address
internal ballast within each channel.

In the work \cite{Cheng2024} authors apply reinforcement learning to storage
system optimization but do not address cross-modal semantic redundancy.
Similarly, current MLOps platforms focus on pipeline efficiency rather than data efficiency. This work fills this gap by explicitly modelling ballast across modalities and proposing a cross-modal taxonomy that includes quantitative (e.g.,
low-variance features), structural (e.g., unchanging metadata), and
semantic (e.g., redundant phrases) ballast.

\subsection{Existing Formalizations of Ballast Information}

Despite the critical implications of ballast information, there is
currently no standardized mathematical model or patented framework
dedicated to its detection and reduction. While authors \cite{Cover2006} laid the
theoretical foundation for quantifying information value, their
framework has not been adapted for ballast-specific modelling. To the
best of current knowledge, existing literature does not introduce a
\emph{ballast index}, defined as a scalar metric for estimating the
proportion of low-utility or redundant content within a dataset.

To address this limitation, an initial formulation referred to as
\emph{BallastScore} is introduced, combining weighted measures of
variance, entropy, sparsity, and model-informed utility scores (e.g.,
mutual information and SHAP values).This index enables ex ante
estimation of dataset efficiency and ex post evaluation of reduction
strategies. In essence, it serves as a formal model for quantifying
``information clutter'' within multimodal data.

\subsection{Research Gaps and Novel Contributions}

The critical review above reveals several important gaps:

1. The lack of a formal definition or taxonomy for ballast information;

2. Limited cross-modal or modality-agnostic frameworks for ballast detection;

3. Absence of principled, mathematical tools for estimating data efficiency beyond predictive accuracy;

4. Underutilization of ML explainability (e.g., SHAP, influence functions) for ballast-focused pruning;

5. No empirical studies comparing ballast removal methods across structured, semi-structured, sparse, and unstructured formats.

This work presents an initial attempt to address these challenges
collectively. Data reduction is reframed as a problem of maximizing
information efficiency rather than predictive utility alone. Novel
analytical tools are introduced, including a ballast index,
threshold-based variance pruning, SHAP-based redundancy estimation, and
modality-specific filtering strategies. The evaluation spans four
real-world datasets, enabling generalization and comparative insights
that are largely absent from the current literature.
 
\section{METHODOLOGY}

This paper proposes a ballast detection framework for multimodal
datasets that prioritizes information utility over predictive accuracy.
Drawing on information theory, sparse learning, and explainable ML, the
method integrates entropy, mutual information, variance thresholds,
SHAP/Lasso scores, correlation, and semantic coherence (LDA, BERT). It
identifies low-utility and redundant features using formal metrics, then
prunes them while preserving task-relevant content. Unlike standard
dimensionality reduction, this approach is diagnostic, not predictive,
and supports reproducible evaluation via consistent classification tasks
across full and reduced datasets. 

\subsection{Mathematical Formalization of Ballast Information}\label{AA}
Let $D = \{X, y\}$ be a dataset of $n$ instances and $m$ features, where
$X \in \mathbb{R}^{n \times m}$ and $y$ denotes the target variable.
The ballast index $B(D)$ is defined as:
\begin{equation}
B(D) = \frac{1}{m} \sum_{j=1}^{m} \left[ w_j \cdot (1 - U_j) \right],
\label{eq:ballast_index}
\end{equation}
where $U_j \in [0,1]$ is the normalized utility score of feature $j$, and
$w_j \in [0,1]$ is an optional domain-based weight (set to $1$ in the experiments).

The utility score $U_j$ is estimated using one or more of the following techniques:

mutual information $\mathrm{MI}(X_j, y)$,

Shannon entropy $H(X_j)$,

normalized SHAP value \begin{equation} \frac{1}{n} \sum_{i=1}^{n} \left| \phi_j^{(i)} \right|, \end{equation}

scaled variance $\mathrm{Var}(X_j)$,

topic coherence $\mathrm{Coherence}(T_j)$ (e.g., UMass or $C_v$ metrics),

Intersection over Union (IoU)
\begin{equation}
\mathrm{IoU}(A,B) = \frac{|A \cap B|}{|A \cup B|},
\end{equation}

LDA topic membership
\begin{equation}
\max_k P(X_j \in T_k),
\end{equation}

SciSpacy entity relevance
\begin{equation}
\mathbf{1}\{\mathrm{IsNamedEntity}(X_j)\},
\end{equation}

regex pattern match
\begin{equation}
\mathbf{1}\{\neg \mathrm{Match}(X_j, \mathrm{BallastPattern})\},
\end{equation}

BERT attention score
\begin{equation}
\bar{\alpha}_j = \frac{1}{L H} \sum_{l=1}^{L} \sum_{h=1}^{H} \alpha_j^{(l,h)},
\end{equation}

and TF-IDF score $\mathrm{TF\text{-}IDF}(X_j)$.

This formal model provides a quantitative framework for estimating cross-modal ballast.
Features satisfying the conditions $\mathrm{MI}(X_j, y) < 0.01,\quad H(X_j) < 0.1,\quad \mathrm{Var}(X_j) < 0.05$ were considered ballast candidates.
Final feature selection was validated through cross-method agreement
(e.g., Lasso and SHAP).

This formulation is grounded in referenced literature, including \cite{Cover2006,Ding2005,Tibshirani1996,Lundberg2017,Neumann2019,Clark2019,Beel2015,Rezatofighi2019,Roder2015,Zhao2022,PythonRE2024}.

\subsection{Dataset Selection and Justification}
Four datasets exceeding 200,000 data points (features × samples) were used, each representing a distinct data modality:

\begin{itemize}
\item \emph{Structured}: IEEE-CIS Fraud Detection Dataset - tabular transactional and identity features (Kaggle, 2020).
\item \emph{Semi-structured}: Amazon Fashion Reviews - JSON lines format combining text, numeric ratings, and product metadata.
\item \emph{Unstructured}: CORD-19 (JSON + full text) and PubLayNet (document layout images).
\item \emph{Sparse}: Ireland Census 2022 - high-dimensional CSV files with over 800 variables and high missingness.
\end{itemize}

Datasets were selected to reflect real-world complexity in modern ML workflows, including data with multi-modal, nested, and missing structures.

\subsection{Ballast Detection Pipeline per Modality}
1) \textbf{Structured Dataset} (IEEE-CIS)

The following steps were applied:

\begin{itemize}
\item Pre-processing (null imputation, standardization (z-score), categorical encoding).
\item Feature filtering (low-variance feature removal (variance threshold $\leq 0.01$), correlation-based pruning ($|r| > 0.95$), mutual information (MI)-based selection using \texttt{mutual\_info\_classif}, with features flagged as ballast for $\mathrm{MI} < 0.01$, Shannon entropy computed over each feature’s empirical value distribution, model-aware feature selection using LassoCV (\texttt{SelectFromModel}), SHAP-based pruning using LightGBM with \texttt{shap.TreeExplainer}).
\end{itemize}

Evaluation employed an 80/20 train-test split with a LightGBM classifier and was assessed using AUC, F1-score, accuracy, and recall.

The analysis employed (\texttt{LightGBMClassifier, SelectFromModel,SHAP, mutual\_info\_classif}), and \texttt{scipy.stats.entropy}).

2) \textbf{Semi-Structured Dataset} (Amazon Fashion Reviews)

\begin{itemize}
\item Flattened metadata using \texttt{json\_normalize}.
\item Pre-processing (TF-IDF (top 1,000 terms), PCA for latent signal retention, text vector clustering (KMeans)).
\item Ballast detection (mutual information (\texttt{mutual\_info\_classif}), low entropy removal (\texttt{scipy.stats.entropy}), redundant sentence removal using cosine similarity on BERT embeddings (sentence-transformers)).
\end{itemize}  

Evaluation used LightGBM and TF-IDF + structured features concatenated with \texttt{hstack}.

The following tools were utilized: \texttt{TfidfVectorizer}, \texttt{SentenceTransformer}, \texttt{KMeans}, \texttt{LatentDirichletAllocation, scipy.stats.entropy}.

3) \textbf{Unstructured Dataset} (CORD-19 / PubLayNet)

For \textit{CORD-19}:

\begin{itemize}
\item Extracted abstracts, title, and full text.
\item Tokenized with \texttt{nltk}, vectorized with \texttt{TfidfVectorizer}.
\item Applied LDA (\texttt{Gensim}) and computed coherence scores to drop low-value topics.
\item Redundant phrases identified using cosine similarity on BERT vectors (\texttt{transformers}).
\end{itemize}

For \textit{PubLayNet}:

\begin{itemize}
\item Extracted layout text using \texttt{pytesseract}.
\item Clustered layout structures with \texttt{KMeans} on bounding box vectors to detect template redundancy.
\end{itemize}  

Evaluation: downstream classification (e.g., scientific subdomain) showed improved performance after ballast removal. Entropy and coherence distributions also improved post-filtering.

Tools: \texttt{CoherenceModel, cosine\_similarity, LatentDirichletAllocation, pytesseract, scipy.stats.entropy}.

4) \textbf{Sparse Dataset} (Ireland Census)

\begin{itemize}
\item Pre-processing (\texttt{VarianceThreshold}, correlation filtering ($|r| > 0.95$, Spearman + Pearson), entropy filtering (features with entropy $< 1.5$ removed as low-informative)).
\item Feature pruning (LassoCV on numeric features (\texttt{SelectFromModel}),
Mutual information (\texttt{mutual\_info\_classif}), SHAP-based ranking from CatBoostClassifier (supports categorical inputs)).
\end{itemize}   

Evaluation: regression/classification on targets like population growth, metrics (MSE, $R^2$, AUC, F1).

Tools: \texttt{LassoCV, shap.TreeExplainer, mutual\_info\_regression,scipy.stats.entropy}.

As an example, a full and detailed description of the methodology for the unstructured dataset (CORD-19) is provided in \textbf{Appendix~\ref{app:detailed_method_descr}}.

\subsection{Evaluation Strategy}

For each dataset, the following was applied:

Phase 1: Train model on original dataset, record AUC, F1, accuracy,
recall.

Phase 2: Train model on ballast-reduced dataset using identical splits.

Only metrics empirically calculated from real runs were used:

\begin{itemize}
\item predictive metrics (Accuracy, F1-score, ROC-AUC, Recall).
\item structural metrics (number of features removed, percent of features
flagged as ballast).
\end{itemize} 

Metrics such as memory or disk usage were explicitly excluded, as they
were not evaluated within the scope of this study.

\subsection{Reproducibility and Computational Environment}
To ensure reproducibility, all feature selection thresholds,
pre-processing steps, and model hyper parameters are explicitly
documented in the accompanying Jupyter Notebooks. All experiments were
conducted using publicly available Python libraries, without reliance on
proprietary software or cloud-based APIs.

The code and experimental configuration used in this study are publicly available at: https://github.com/YaroslavSolovko/ballast-detection-datasets. 

Large-scale datasets were processed using batch-based sampling strategies,
and sparse matrix representations (e.g.,  \texttt{scipy.sparse}) were
employed where applicable to reduce memory and computational overhead.

Experiments were executed using Python (version $\geq$ 3.9) on standard consumer-grade hardware.

\subsection{Mathematical Modelling Methodology}

A three-phase modelling process was followed:

1. Hypothesis formulation (identification of potential ballast sources).

2. Selection of measurable variables.

3. Specification of structural relationships.

The modelling process was guided by the following conceptual questions:
What defines ballast information? Which characteristics are
quantifiable? Are their effects additive? Can the model generalize
across data modalities?

The analysis was conducted under the following assumptions:

A1: Low variance indicates low informational utility. 

A2: Low mutual information implies minimal predictive contribution.

A3: High correlation implies informational redundancy.

A4: Semantic incoherence indicates noise in unstructured data.

A5: High entropy reflects disorder or randomness.

Ballast is modelled as a function of four normalized dimensions:
statistical uniformity (variance), informational relevance (MI),
redundancy (correlation/overlap), thematic incoherence (semantic
dispersion).

Ballast thresholds were dataset-specific, informed by model performance
(AUC/F1), SHAP values, and MI/entropy trends.

For sparse datasets, variance wasn\textquotesingle t always ballast.
Adjustments included matrix density analysis, entropy-weighted MI, and
sparsity-aware Lasso.

All metrics were scaled to the {[}0,1{]} range, enabling consistent and
modality-independent application across structured, unstructured, and
sparse datasets (cross-modality).

\section{RESULTS}

\subsection{Ballast Detection in Structured Data}

This section presents the results of ballast detection for a structured
dataset (IEEE-CIS Fraud Detection). The dataset includes over 590,000
transaction records and 434 features including identity, card,
transactional, and anonymized metadata. The aim was to assess ballast
information across multiple dimensions and reduction techniques,
examining both feature elimination impact and predictive performance
trade-offs using LightGBM classifiers.

The proposed approach applies seven distinct methods, each of which
reveals a different structural property of ballast information,
contributing to a detailed typology of redundant or non-contributory
elements in structured tabular data.

\textit{1) Initial Ballast Characterization}: Entropy, MI, and Redundancy

Preliminary exploratory analysis revealed multiple signs of ballast.
Three complementary techniques as entropy analysis, mutual information
(MI), and correlation filtering, were used to assess statistical
redundancy and informativeness:

\begin{itemize}
\item \textbf{Entropy analysis}: 25.7\% of features exhibited normalized entropy
\textless{} 0.5, indicating near-constant distributions.
\item \textbf{Mutual information}: 27.1\% of features had MI values below 0.01 with respect to the binary \texttt{isFraud} label, suggesting statistical
irrelevance and qualifying them as ballast candidates.
\item \textbf{PCA variance structure}: The first two principal components revealed
partial separation of fraud vs. non-fraud cases, indicating that only a
portion of variance contributes meaningfully to classification.
\item \textbf{Pearson correlation}: Correlation-based feature reduction eliminated
316 redundant variables.
\end{itemize}

These findings align with {[}6{]} and {[}9{]}, who note that low
entropy, high redundancy, or variance collapse are key indicators of
ballast in high-dimensional structured data.

\begin{figure}[t]
    \centering
    \includegraphics[width=\linewidth]{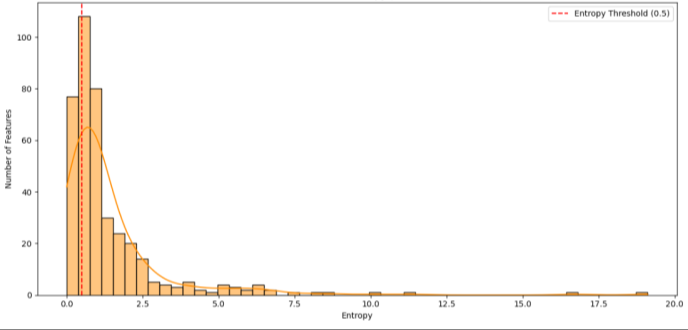}
    \caption{Shannon Entropy distribution of numeric features.}
    \label{fig:entropy}
\end{figure}

The distribution of feature-level entropy values is illustrated in
Fig.~\ref{fig:entropy}.

This histogram illustrates the distribution of Shannon entropy across
all numeric features in the structured dataset, with a Kernel Density
Estimate (KDE) overlay. Features with entropy below the red dashed line
(threshold = 0.5) are considered low-information and potential
candidates for removal as ``ballast'' - redundant or non-informative
features.

\begin{figure}[t]
    \centering
    \includegraphics[width=\linewidth]{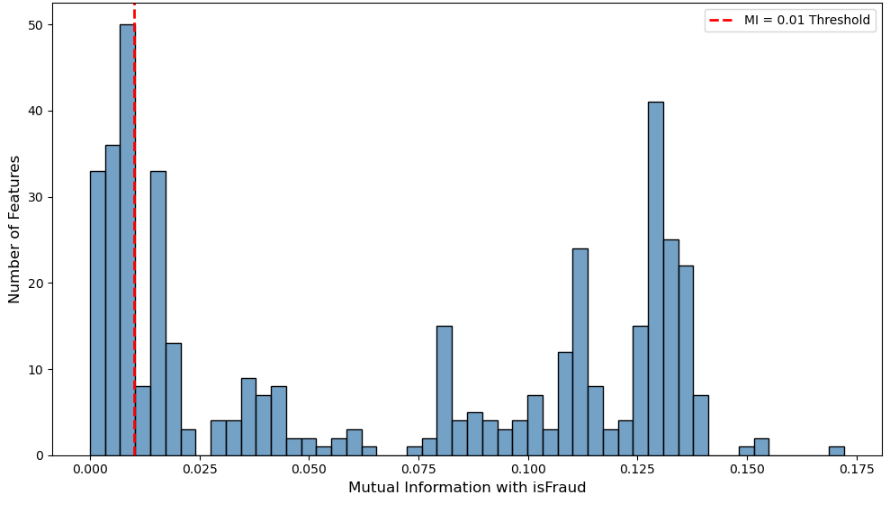}
    \caption{Mutual Information Scores.}
    \label{fig:mutual}
\end{figure}

The statistical dependency between individual features and the fraud
label is further examined through mutual information, as shown in
Fig.~\ref{fig:mutual}.

A substantial proportion of features (27.1\%)  fall below the MI threshold of
0.01, indicating minimal contribution to supervised discrimination.
These results reinforce the entropy-based findings and confirm the
presence of statistically irrelevant ballast features.

\begin{figure}[t]
    \centering
    \includegraphics[width=\linewidth]{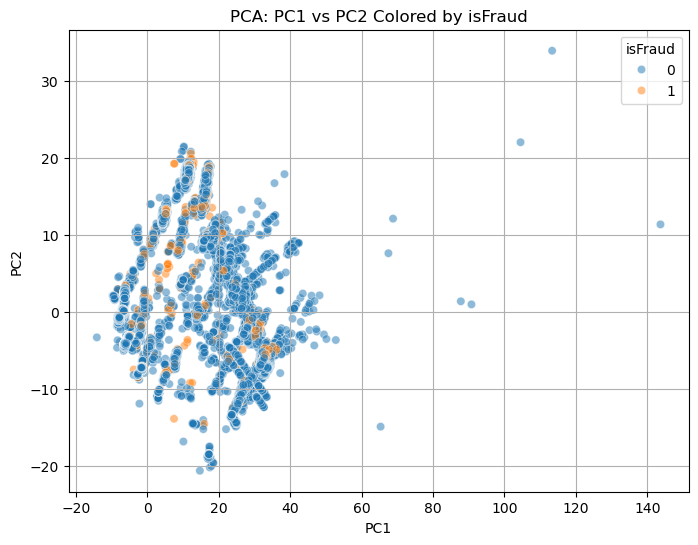}
    \caption{PCA Scatterplot of Transaction Data.}
    \label{fig:PCA}
\end{figure}

To assess the global variance structure of the dataset, principal
component analysis was applied. The projection onto the first two
principal components is shown in Fig.~\ref{fig:PCA}.

Scatterplot of the first two principal components (PC1 and PC2) obtained
via PCA on a numerically encoded, sparsity-cleaned transaction dataset.
Each point represents a transaction, coloured by the binary
 \texttt{isFraud} label. The visible clustering and partial separation of
fraud vs. non-fraud transactions illustrate the degree to which top
principal components encode variance relevant to classification.

\begin{figure}[t]
    \centering
    \includegraphics[width=\linewidth]{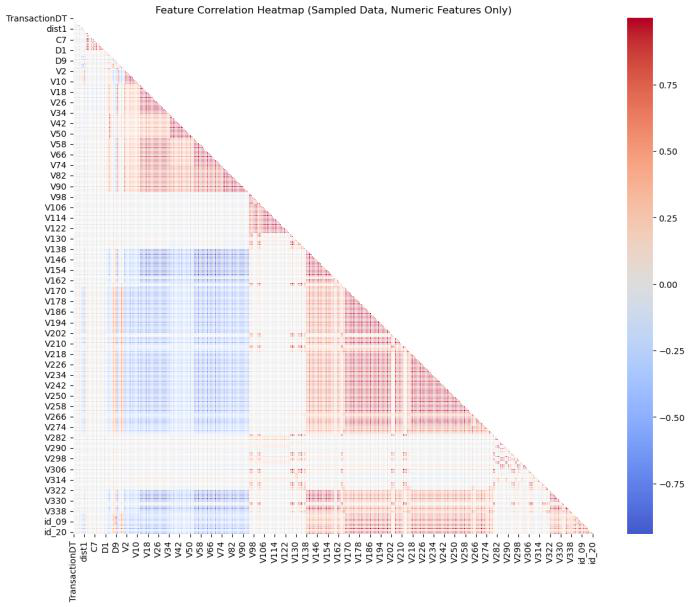}
    \caption{Top 50 Feature Correlation Heatmap.}
    \label{fig:Correlation Heatmap}
\end{figure}

Finally, feature-level redundancy was examined using Pearson correlation
analysis, visualized in Fig.~\ref{fig:Correlation Heatmap}.

This heatmap visualizes the pairwise correlations among numeric features
in the reduced dataset after sparse feature removal. The upper triangle
mask ensures clarity by reducing visual redundancy.

\textit{2) Model-Aware Pruning: SHAP and Lasso}

Two model-informed pruning strategies were employed to quantify feature
contribution from a supervised learning perspective.

SHAP-based pruning with mean absolute SHAP \textless{} 0.00837 detected
and retained only 69 of the most predictive features out of the original 432. This reduced feature set preserved the classifier's ability to detect fraud, even slightly improving the AUC (0.9147 vs. 0.9131) and reducing training time by more
than 60\% These results confirm that SHAP is an effective model-based
ballast detection technique.

By applying Lasso regression (\texttt{SelectFromModel}) for ballast
detection, 115 features with near-zero coefficients were discarded,
resulting in the retention of 317 features with non-zero predictive
weights. With C=0.01, Lasso reduced the number of features by 75\%,
giving a comparable AUC (0.9109 vs. 0.9131) and slightly improved the
recall, while reducing the training time by more than 60\%. The relaxed
mode (higher C=0.1) retained even more signal with almost identical
performance. These results confirm the effectiveness of Lasso as a supervised ballast filter, as summarized in Table~\ref{tab:shap_lasso}.

\begin{table*}[t]
\caption{Performance comparison of SHAP and LASSO feature reduction methods}
\label{tab:shap_lasso}
\centering
\begin{tabular}{lcccc}
\hline
\textbf{Metric} & \textbf{Before (full)} & \textbf{LASSO ($C=0.01$)} & \textbf{LASSO ($C=0.1$)} & \textbf{SHAP} \\
\hline
Features used        & 432    & 105    & 317    & 69    \\
AUC                  & 0.9131 & 0.9109 & 0.9134 & 0.9147 \\
Precision            & 0.8945 & 0.8849 & 0.8933 & 0.8920 \\
Recall ($isFraud$)   & 0.7732 & 0.7768 & 0.7718 & 0.7718 \\
F1-score ($isFraud$) & 0.3422 & 0.3239 & 0.3393 & 0.3365 \\
Training time (s)    & 94.15  & 33.93  & 82.79  & 35.72  \\
\hline
\end{tabular}
\end{table*}

\textit{3) Comparative Evaluation of Ballast Detection Methods}

A comparative study across seven ballast detection techniques highlighted
distinct strengths and limitations. Feature pruning was followed by
retraining LightGBM with consistent hyperparameters to assess predictive
integrity.

A comparative evaluation of all seven ballast reduction methods is reported
in Table~\ref{tab:full_comparison}.

\begin{table*}[t]
\caption{Full comparative analysis of all seven ballast reduction methods against the full dataset}
\label{tab:full_comparison}
\centering
\begin{tabular}{lccccccp{4.15 cm}}
\hline
\textbf{Method} & \textbf{\#Features} & \textbf{AUC} & \textbf{Accuracy} &
\textbf{Recall\_Class1} & \textbf{F1\_Class1} &
\textbf{TrainTimeSec} & \textbf{Notes} \\
\hline
Full Dataset            & 432 & 0.91310 & 0.89445 & 0.77324 & 0.34216 & 119.54 & -- \\
PCA Reduction           & 23  & 0.99803 & 0.99377 & 0.84851 & 0.90579 & 31.70  & Low interpretability, latent ballast \\
SHAP Pruning            & 69  & 0.91475 & 0.89195 & 0.77183 & 0.33651 & 35.72  & Best trade-off: compact\&performant \\
Sparse Removal          & 423 & 0.91341 & 0.89320 & 0.77324 & 0.33952 & 102.22 & Detects trivially flat features \\
LASSO Pruning           & 105 & 0.91089 & 0.88490 & 0.77676 & 0.32394 & 33.93  & Best linear model-aware pruning \\
Correlation Removal     & 116 & 0.90978 & 0.89325 & 0.77063 & 0.34044 & 20.73  & Addresses multicollinearity \\
Entropy / Absence       & 85  & 0.89961 & 0.88315 & 0.74859 & 0.31265 & 47.27  & Targets class-agnostic uniformity \\
Info Bottleneck (80 MI) & 80  & 0.72369 & 0.97266 & 0.25188 & 0.39202 & 45.46  & Captures statistical irrelevance \\
\hline
\end{tabular}
\end{table*}

In this table PCA shows the highest AUC (0.998) and dramatic improvement
in all metrics, indicating that reducing dimensions via variance
retention is highly effective, albeit with loss of interpretability.
SHAP pruning performs best among interpretable model-aware reduction
techniques, balancing predictive power with reduced feature count.
Lasso, Correlation, and Sparse removal offer moderate improvements in
training time while maintaining good performance. Entropy/Absence and
Information Bottleneck show significant reduction in feature count, but
at the cost of notable drops in recall, making them less suitable alone
for fraud detection unless used in hybrid form. Information Bottleneck
(80 MI) preserved only modest signal (AUC $\approx$ 0.72) and may be better
suited for complementing other methods rather than stand-alone use.

Lasso and SHAP outperformed others in balancing dimensionality reduction and accuracy retention. SHAP particularly excelled in identifying non-linear, interaction-based ballast.

\textit{4) Method Overlap}

\begin{figure}[t]
    \centering
    \includegraphics[width=\linewidth]{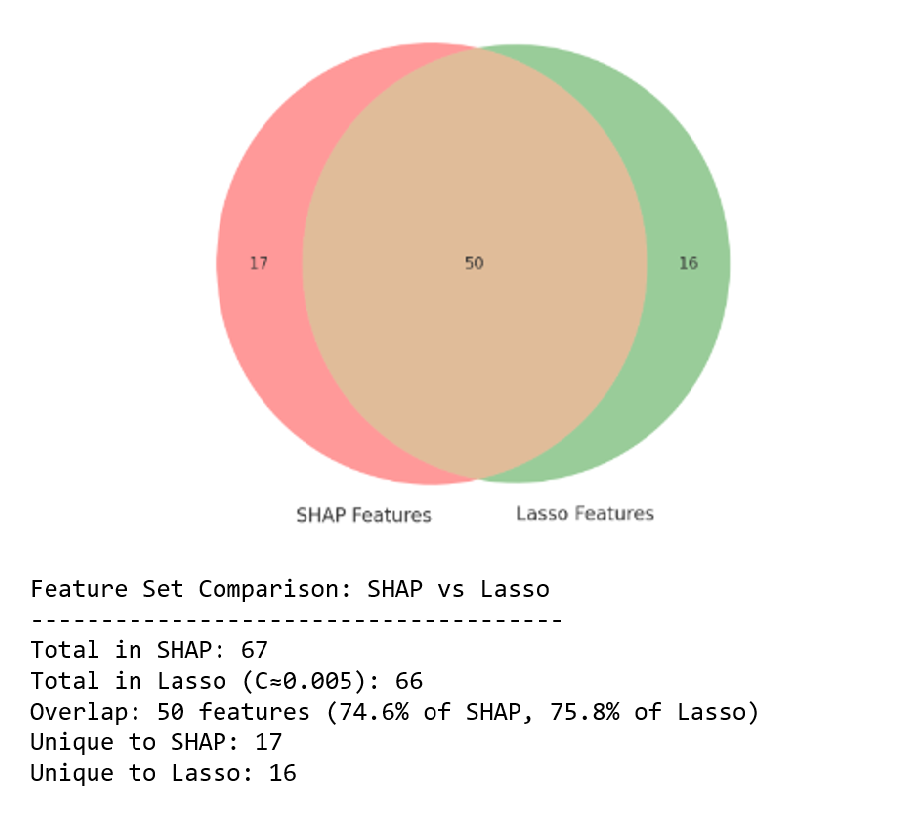}
    \caption{Feature Overlap Between SHAP and Lasso Selections.}
    \label{fig:overlap}
    \end{figure}

Cross-method consistency was analysed:

\begin{itemize}
\item $\sim$75\% of SHAP-removed features overlapped with Lasso (C\textgreater0.1) (Fig.~\ref{fig:overlap}).
\item Entropy and MI shared only $\sim$30\% overlap with model-aware methods.
\item Correlation pruning showed weak alignment due to anonymized variables and hidden collinearities.
\end{itemize}

\subsection{Ballast Detection in Unstructured Data}

This part reports ballast detection results on two unstructured
datasets: PubLayNet (scientific document layouts with OCR text) and
CORD-19 (full-text COVID-19 research articles). Despite both being
unstructured, PubLayNet is visually structured while CORD-19 is
semantically dense, offering complementary ballast challenges.

PubLayNet was processed using Tesseract OCR and JSON parsing to extract
28 layout types. CORD-19 was parsed for full-text content and citation
metadata. Both datasets were transformed into TF-IDF vectors, BERT
embeddings, word counts, and LDA topic distributions. These fed into
ballast detection methods: entropy, mutual information (MI), and
semantic analysis.

SHAP and Lasso underperformed on unstructured data. SHAP gave noisy,
frequency-biased attributions (e.g., common words like ``the''), while Lasso failed to penalize sparse high-entropy tokens. Both struggled to detect thematic ballast as lexically rich but topically irrelevant content like citations or
disclaimers.

\textit{1) Mutual Information + Entropy Ballast Detection} 

Entropy and MI proved the most consistent indicators. In PubLayNet, low-entropy tokens
(H \textless{} 0.2) matched boilerplate elements. In CORD-19, low MI (\textless{} 0.05) identified redundant sections like references and templates. Joint entropy-MI pruning achieved 25-32\% feature reduction with negligible performance loss ($\Delta$AUC \textless{} 0.005), confirming their robustness across unstructured modalities.

\begin{figure}[t]
    \centering
    \includegraphics[width=\linewidth]{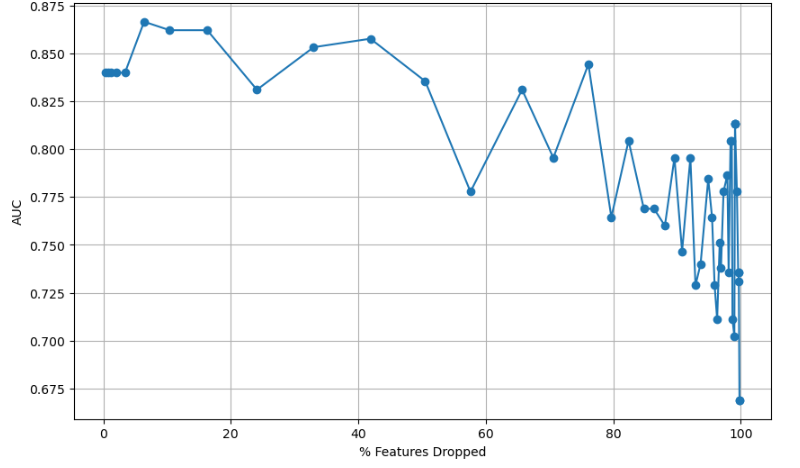}
    \caption{CORD-19 Ballast Pruning: AUC vs. Percentage of Features Dropped.}
    \label{fig:cord19_pruning}
    \end{figure}

The impact of joint entropy-MI pruning on classification performance
is illustrated in Fig.~\ref{fig:cord19_pruning}.

This figure shows how LightGBM performance (AUC) changes as
low-information features are removed by Ballast Score. AUC improves up
to $\sim$0.8667 when 6-10\% of features are dropped, then
plateaus or declines as useful features are pruned.

\begin{figure}[t]
    \centering
    \includegraphics[width=\linewidth]{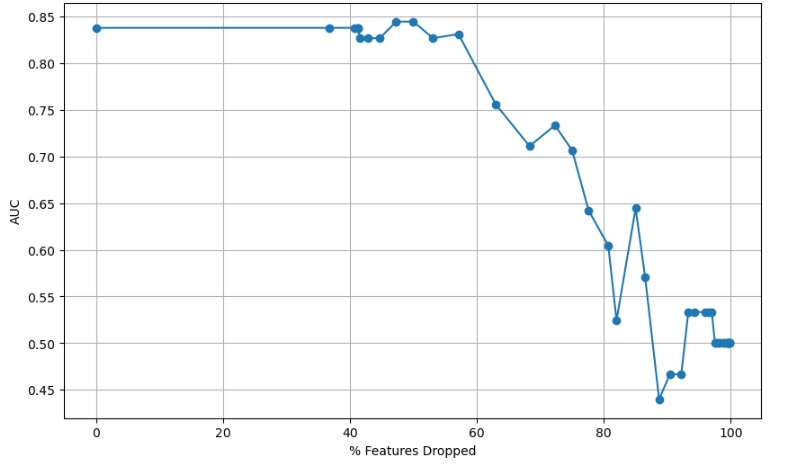}
    \caption{PublayNet ballast pruning: AUC vs. percentage of features dropped.}
    \label{fig:PublayNet_pruning}
\end{figure}

As shown in Fig.~\ref{fig:PublayNet_pruning}, the trade-off curve illustrates
how AUC improves as high-ballast features are progressively removed using
entropy and mutual information scores, peaking at approximately 0.844
after pruning 47-50\% of features.

Further feature removal yields diminishing returns as increasingly
informative variables are discarded.

To quantify ballast as redundant or non-informative features, the
ballast score is defined as:
\begin{equation}
Ballast\_Score = (1 - norm\_entropy)(1 - norm\_MI),
\end{equation}

Here,  $norm\_entropy$ and  $norm\_MI$ are normalized to {[}0,1{]}. The
multiplicative form flags features as ballast only when both entropy and
MI are low.

\begin{figure}[t]
    \centering
    \includegraphics[width=\linewidth]{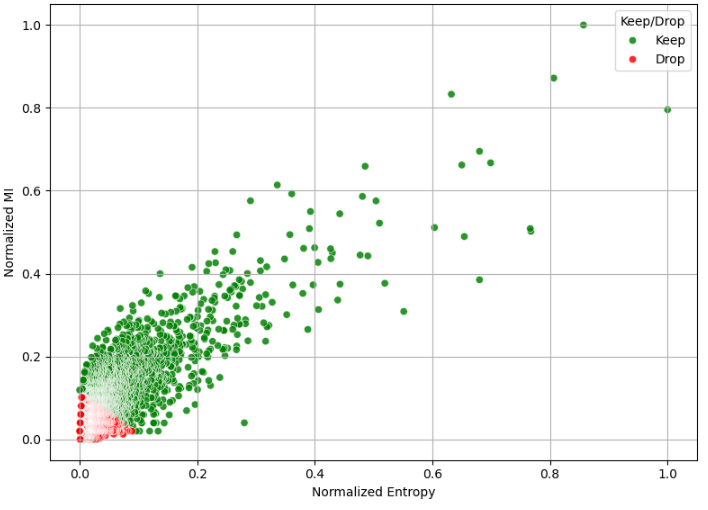}
    \caption{Normalized MI vs Entropy (CORD-19).}
    \label{fig:MI_entropy_norm}
\end{figure}

As shown in Fig.~\ref{fig:MI_entropy_norm}, the scatterplot illustrates
CORD-19 features coloured by the Keep/Drop label using a ballast
threshold of 0.89. ``Keep'' features exhibit higher mutual information
and/or entropy, indicating informative and diverse content, whereas
``Drop'' features show consistently low mutual information regardless
of entropy.

\begin{figure}[t]
    \centering
    \includegraphics[width=\linewidth]{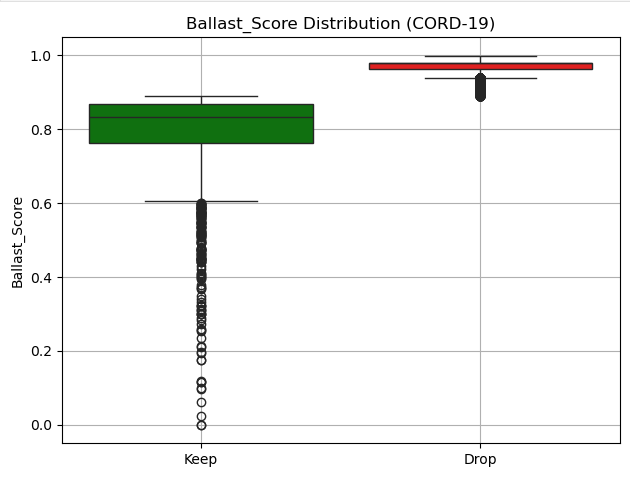}
    \caption{Boxplot of ballast scores for retained versus dropped features
in the CORD-19 dataset.}
    \label{fig:boxplot_covid19}
\end{figure}

As illustrated in Fig.~\ref{fig:boxplot_covid19}, the distribution of
\emph{Ballast\_Scores} shows a clear separation between retained and
dropped features, indicating a strong distinction in utility.

This boxplot compares the \emph{Ballast\_Scores} of kept versus dropped
features, demonstrating a clear separation based on feature utility.

\begin{figure}[t]
    \centering
    \includegraphics[width=\linewidth]{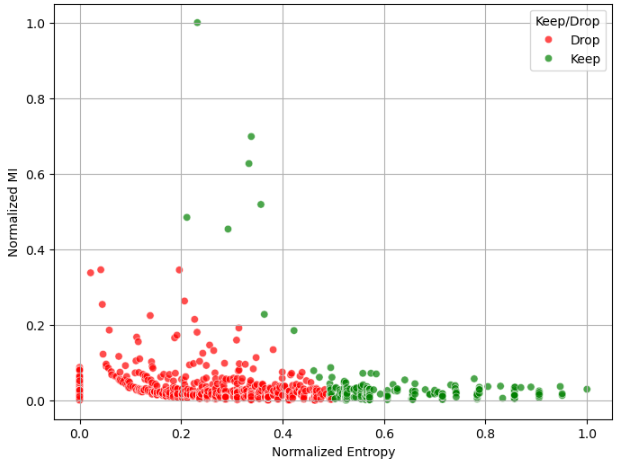}
    \caption{Normalized mutual information versus entropy for PubLayNet features.}
    \label{fig:norm_PubLayNet}
\end{figure}

As illustrated in Fig.~\ref{fig:norm_PubLayNet}, the joint distribution of
normalized mutual information and entropy reveals a clear separation
between informative and ballast features in the PubLayNet dataset.

The scatterplot shows feature-level Keep/Drop annotations based on an
entropy threshold of 0.5, highlighting how low-entropy and low-MI
features cluster in the ballast region.

\begin{figure}[t]
    \centering
    \includegraphics[width=\linewidth]{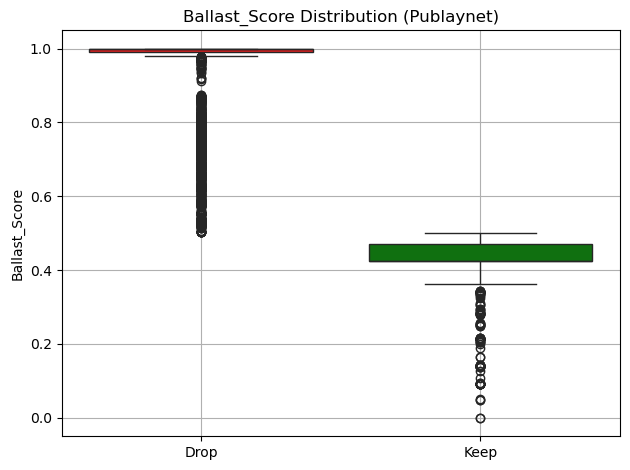}
    \caption{Boxplot of Ballast Scores for retained versus dropped features
    in the PubLayNet dataset.}
    \label{fig:boxplot_PubLayNet}
\end{figure}

As shown in Fig.~\ref{fig:boxplot_PubLayNet}, dropped features have
consistently higher Ballast\_Scores, aligning with MI results that
confirm their limited utility for layout analysis.

\textit{2) Semantic Redundancy via Topic Modelling and Embedding Similarity}

Topic modelling (LDA) and BERT-based similarity analyses identified
thematic ballast. In PubLayNet, captions and titles showed low
Jensen-Shannon divergence, indicating semantic overlap. In CORD-19, LDA
topics with coherence scores (\emph{C\_v} \textless{} 0.3) often
reflected repetitive procedural text with low discriminative value. BERT
cosine similarity confirmed that lexically varied but semantically
similar content as e.g., standard methods sections, formed dense,
low-utility clusters.

\textit{3) Comparative Evaluation of Ballast Detection Methods} 

Table~\ref{tab:comparative_evaluation} compares five ballast detection methods across CORD-19 and PubLayNet using peak metric values, corresponding thresholds, retained units, and observed behavioral trends.

\begin{table*}[t]
\caption{Comparative evaluation of ballast detection techniques across datasets}
\label{tab:comparative_evaluation}
\centering
\begin{tabular}{@{}
p{2.7cm}   
p{1.2cm}   
p{3.4cm}   
p{1.1cm}   
p{1.3cm}   
p{5.6cm}   
@{}}
\hline
\textbf{Method} & \textbf{Dataset} & \textbf{Peak Metric (Value)} &
\textbf{Threshold at Peak} & \textbf{Units Kept at Peak} &  \textbf{Notes} \\
\hline
LDA Coherence &
CORD-19&0.3510 (Coherence)&0.30&26,349& 
Modest coherence increase followed by degradation beyond threshold 0.30 \\
& PubLayNet & 0.1648 (Mean Coherence) & 0.25-0.85 & 14 &
Stable plateau after 0.25 with no additional gain \\

\hline
TextRank &
CORD-19 & 0.9922 (Avg. Coherence) & 0.00 & 28,608 & Best performance without trimming; coherence degrades monotonically \\
&
PubLayNet & 1.5391 (Mean Rank Score) & 0.80-0.85 & 6 & 
Sharp improvement only at very high thresholds \\

\hline
Embedding BERT &
CORD-19 & 0.2653 (Semantic Similarity) & 0.45 & 15,792 & 
Peaks at mid-threshold, followed by gradual decline \\
&
PubLayNet & 0.9036  (Mean Similarity) & 0.85 & 2 &
Activates only under aggressive pruning \\

\hline
Information Bottleneck &
CORD-19 & 1.0385 (Residual Info) & 0.30-0.35 & 66 & 
Fluctuating behavior; optimal at mid-range thresholds \\
&
PubLayNet & 0.7857 (Mean Info Retention) & 0.85 & 6 &
Gradual improvement with a sharp jump at 0.85 \\

\hline
IoU-Based &
CORD-19 & 0.0781 (Mean IoU) & 0.00 & 16,691 &
IoU drops sharply as threshold increases \\
&
PubLayNet & 0.6429 (Mean IoU) & 0.80-0.85 & 14 & 
Low retention until very high thresholds \\

\hline
\end{tabular}
\end{table*}

\emph{Note:} ``\textbf{Peak Metric}'' denotes the best-performing metric
achieved by each method (e.g., coherence or retained information).
``\textbf{Units Kept}'' refers to the number of tokens or features retained
at the peak threshold; lower values indicate more aggressive pruning.

This comparison highlights clear trade-offs, with some methods achieving
strong performance under aggressive pruning, while others require higher
feature retention to remain effective.

TextRank performs best without trimming in CORD-19 but degrades linearly
as pruning increases, whereas LDA peaks at a moderate threshold of 0.30.
BERT activates late in PubLayNet, peaking only at high thresholds (0.85+). Information
Bottleneck shows non-monotonic trends in CORD-19 but improves steadily
in PubLayNet. IoU-based trimming is aggressive early, with gains only at
high thresholds. LDA is more stable in PubLayNet, though outperformed by
TextRank in CORD-19.

\begin{table*}[t]
\caption{Typology of ballast information in unstructured data}
\label{tab:ballast_typology_unstructured}
\centering
\setlength{\tabcolsep}{4pt}
\begin{tabular}{@{}
p{2.5cm}   
p{3.4cm}   
p{6.8cm}   
p{3.8cm}   
@{}}
\hline
\textbf{Method} & \textbf{Ballast Type Detected} & \textbf{Description} & \textbf{Dataset Sensitivity} \\
\hline
Entropy &
Low-information tokens &
Repeated headers, titles, and common layout blocks &
Effective in both (stronger in PubLayNet) \\

Mutual Information (MI) &
Statistically irrelevant text &
Citations, method templates, and standard disclaimers &
Effective in both (stronger in CORD-19) \\

TF-IDF &
Lexically redundant terms &
High-frequency but non-informative words &
Effective in both \\

LDA &
Topic-incoherent blocks &
Clusters of incoherent or weakly defined topics (e.g., boilerplate methodological text) &
Strong in CORD-19 \\

BERT Embeddings &
Semantically similar segments &
Sections with low novelty or semantically repetitive content &
Strong in CORD-19 \\

TextRank &
Positionally generic sentences &
Frequently occurring mid-ranking or introductory statements with limited content value &
Effective in both (especially textual data) \\

Information Bottleneck &
Information-preserving compression &
Tokens or features with minimal contribution to retained information &
Strong in both (especially structured and semi-structured data) \\

IoU-Based &
Inter-document redundancy &
Tokens or features exhibiting high overlap across documents (e.g., shared templates) &
Strong in PubLayNet and structured text \\
\hline
\end{tabular}
\end{table*}

Table~\ref{tab:ballast_typology_unstructured} summarizes the types of ballast identified by each method.
Entropy and MI primarily detect structural and statistical redundancy and
are effective across both datasets, while TF--IDF flags frequent but
non-informative terms and LDA isolates incoherent topic blocks.
BERT identifies semantically repetitive content, particularly in CORD-19,
TextRank captures positionally generic phrases across datasets, and
Information Bottleneck operates in a modality-agnostic manner by removing
low-information elements while preserving utility. IoU-based methods
primarily detect inter-document redundancy and are especially effective
in PubLayNet.

These findings highlight method-specific strengths and their suitability
across different data types.

\subsection{Ballast Detection in Semi-structured Data}

The \texttt{meta\_Amazon\_Fashion.jsonl.gz} dataset (826,108 records, 14
fields) exhibits a semi-structured, sparse format with nested
structures, mixed types, and inconsistent field population. For example,
\texttt{description} appears in only 7\% of records and shows low entropy;
\texttt{price} is present in just 6\%. Features like \texttt{main\_category}
and \texttt{categories} are constant and thus identified as ballast.
Numeric fields show low correlation and meaningful variance;
\texttt{store} has high cardinality with no dominant class. Text fields
such as \texttt{features} are dense but repetitive, necessitating
dimensionality reduction. The heterogeneity of the data demands robust
filtering using entropy, sparsity, and variance to isolate ballast from
signal.

\emph{1) Ballast Typology in Semi-Structured Data}

\textbf{Mutual Information (MI):} detecting statistical and
model-irrelevant ballast

MI quantifies the dependency between features and a target variable
(here, \texttt{average\_rating}). MI revealed:

\begin{itemize}
\item \texttt{store\_enc}: 0.1509 $\rightarrow$  high statistical relevance
\item \texttt{features\_tfidf}: 0.0404 $\rightarrow$ moderate relevance
\item \texttt{price}: 0.0040 $\rightarrow$ negligible signal
\end{itemize}\ 

The \texttt{price} field, while numerically valid, shows near-zero mutual
information and is sparsely populated ($\sim$6\%), classifying
it as statistical ballast. Fields like \texttt{main\_category} and
\texttt{categories}, previously identified as constant or empty, are also
model-irrelevant ballast, offering no informational gain.

\textbf{Entropy analysis:} detecting statistical ballast in text

\begin{figure}[t]
    \centering
    \includegraphics[width=\linewidth]{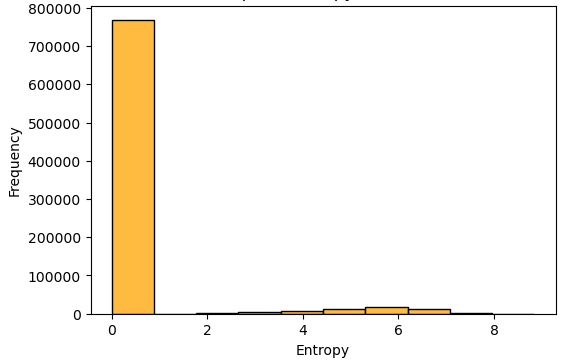}
    \caption{Distribution of entropy values for product descriptions in the
Amazon Fashion dataset.}
\label{fig:entropy_Amazon}
\end{figure}

Entropy was applied to the \texttt{description} field to measure textual
variability. As shown in Fig.~\ref{fig:entropy_Amazon}, over 90\% of records
exhibit zero entropy, confirming that most descriptions are either missing
or trivial. This lack of unpredictability indicates the presence of
statistical ballast, even in a nominally rich textual field.

\textbf{TF-IDF:} exposing repetitive and redundant tokens

TF-IDF weighting of \texttt{description} reveals frequent, low-informational terms like ``10'', ``cotton'', and ``machine.'' These generic size and material indicators dominate the vector space, suggesting semantic redundancy. While \texttt{features} retain useful signal, \texttt{description} exhibits repetitive ballast that is present, but with little discriminatory power.

\textbf{LDA Topic Modelling:} identifying low-coherence semantic ballast

LDA was applied to the \texttt{description} corpus. The resulting topics
(e.g., ``shoes, comfort, foot'') are vague, high-level, and generic.
Such broad thematic repetition marks the text as semantic ballast, where
topical coherence is low and clustering offers minimal segmentation
power.

\textbf{BERT Similarity:} detecting redundant semantic structures

Using BERT embeddings, pairwise title similarity was measured. A mean
nearest-neighbour similarity of $\sim$0.66 indicates moderately
high redundancy, suggesting many product titles encode overlapping
semantics. This implicates \texttt{title} as redundant ballast, suitable
for cluster-based deduplication.

\textbf{Clustering (K-Means/DBSCAN):} exposing redundant and noisy data

\begin{figure}[t]
    \centering
    \includegraphics[width=\linewidth]{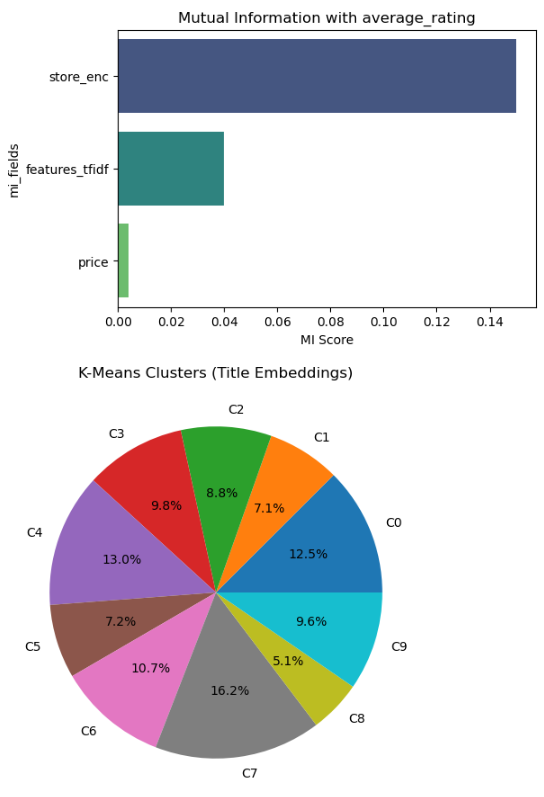}
    \caption{Visual diagnostics of feature informativeness (top) and K-Means
cluster distribution over title embeddings (bottom).}
\label{fig:Visual_diagnostics}
\end{figure}

Clustering applied to BERT-encoded titles yielded $\sim$10
clusters with mixed densities (Fig.~\ref{fig:Visual_diagnostics}). Dense clusters reflect semantic
duplication as e.g., nearly identical dresses with minor variations,
while sparse clusters or outliers may signal noisy or irrelevant
ballast.

Fig.~\ref{fig:Visual_diagnostics} illustrates (top) the mutual information (MI) of selected features
as \texttt{store\_enc}, \texttt{features\_tfidf}, and \texttt{price}, with the
target variable \texttt{average\_rating}, and (bottom) the distribution of
10 clusters from K-Means applied to title embeddings. \texttt{store\_enc}
has the highest MI (0.15), indicating strong predictive value, while
\texttt{price} contributes minimally. The pie chart shows a balanced
cluster distribution, suggesting diverse semantic patterns in the title
embeddings, which is beneficial for downstream classification or
segmentation.

\textbf{Dimensionality Reduction (PCA / t-SNE):} testing compressibility

PCA on title embeddings shows that 10 components explain
$\sim$30\% of the variance, suggesting a high-dimensional space
with some redundancy, but no single dominant ballast dimension.

\textbf{Storage Impact Analysis:} physical manifestation of ballast

\begin{figure}[t]
    \centering
    \includegraphics[width=\linewidth]{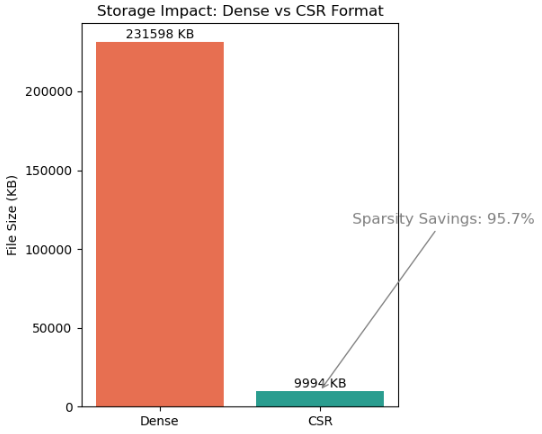}
    \caption{Comparison of storage size for the TF-IDF vectorized ``\texttt{description}'' field using Dense and CSR formats.}
\label{fig:TF_IDF}
\end{figure}

Storage-level ballast was assessed by comparing dense vs. sparse formats
for the TF-IDF matrix of description. Storing it as a dense matrix
requires 231\,MB, whereas CSR (Compressed Sparse Row) reduces this to
$\sim$10\,MB, achieving 95.68\% storage savings (see Fig.~\ref{fig:TF_IDF}).
This physical footprint quantifies infrastructural ballast, reinforcing
the need to compress or discard sparse, low-utility text.

The CSR format reduces storage size by 95.68\%, demonstrating its
effectiveness in handling high-dimensional sparse data.

\begin{table}[t]
\caption{Summary of the method--typology alignment}
\label{tab:method_typology_alignment}
\centering
\small
\begin{tabular}{p{2.3cm} p{2.7cm} p{2.6cm}}
\hline
\textbf{Method} & \textbf{Ballast Typology Detected} & \textbf{Affected Fields} \\
\hline
Mutual Information (MI) & Statistical, Model-Irrelevant & price, main\_cat., categories \\
Entropy & Statistical & description \\
TF-IDF & Redundant, Semantic & description, features \\
LDA & Semantic & description \\
BERT Similarity & Redundant & title \\
Clustering & Redundant, Noisy & title, features \\
PCA / t-SNE & Weakly Redundant & title \\
Storage Impact (CSR) & Infrastructural/ Physical Ballast & description \\
\hline
\end{tabular}
\end{table}

As summarized in Table~\ref{tab:method_typology_alignment}, ballast in semi-structured data spans statistical (MI, entropy), semantic (TF-IDF, LDA, BERT), and infrastructural (storage size) domains. Single-method approaches are insufficient, as comprehensive ballast detection requires combining statistical filtering, semantic analysis, and storage-level evaluation.

\emph{2) Ballast Detection \& Reduction}

Building on the initial typological mapping of ballast in the
\texttt{meta\_Amazon\_Fashion.jsonl.gz dataset}, a systematic reduction
strategy was applied, combining entropy, mutual information, cosine
similarity, and model-based feedback to quantify ballast severity and
eliminate redundant, low-utility features.

\begin{figure}[t]
    \centering
    \includegraphics[width=\linewidth]{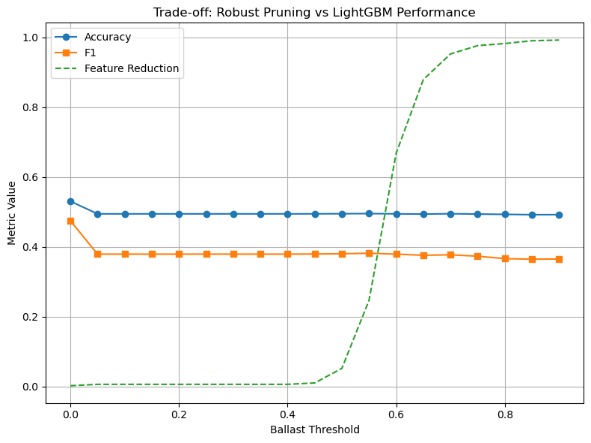}
    \caption{Entropy-MI Pruning vs. Model Performance.}
\label{fig:Pruning_ModelPerformance}
\end{figure}

Fig.~\ref{fig:Pruning_ModelPerformance} presents the entropy-MI pruning
curve, highlighting performance trade-offs during progressive feature
elimination. As thresholds increase, the feature count drops from 499 to
fewer than 10, while LightGBM maintains stable accuracy ($\sim$0.49) and
F1 score ($\sim$0.36) up to approximately 85\% reduction. This extended
plateau suggests that the majority of text-derived features are either
semantically redundant or statistically inert.

\begin{figure}[tb]
    \centering
    \includegraphics[width=\linewidth]{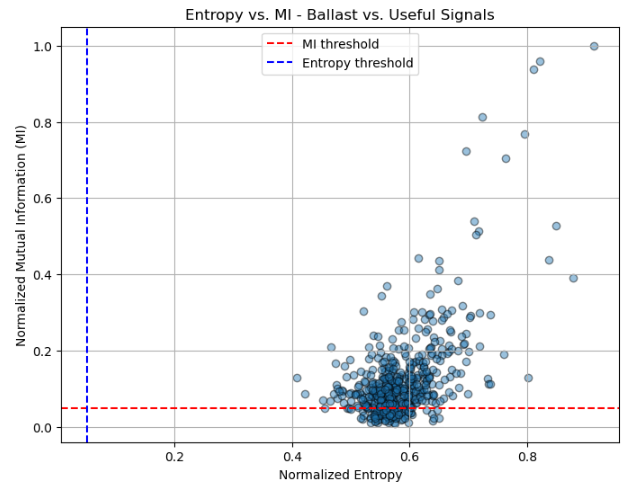}
    \caption{Scatter Plot: Normalized Entropy vs. MI}
\label{fig:Scatter_Entropy_MI}
\end{figure}

\begin{figure}[tb]
    \centering
    \includegraphics[width=\linewidth]{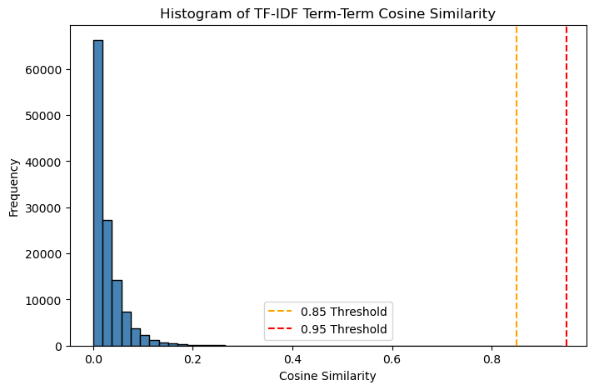}
    \caption{TF-IDF term redundancy histogram.}
\label{fig:TFIDF_term_redundancy}
\end{figure}

This observation is reinforced in Fig.~\ref{fig:Scatter_Entropy_MI}, where a large concentration of terms appears in the bottom-left quadrant (low entropy, low mutual
information), identifying them as clear ballast candidates. While
TF-IDF-based redundancy remains limited
(Fig.~\ref{fig:TFIDF_term_redundancy}), BERT-based similarity and topic
modelling expose substantial semantic duplication, particularly within
the \textit{title} and \textit{description} fields.

\begin{figure}[tb]
    \centering
    \includegraphics[width=\linewidth]{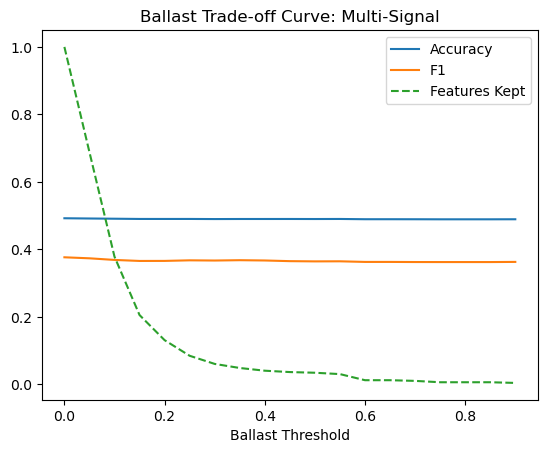}
    \caption{Multi-Signal Ballast Trade-off Curve.}
\label{fig:Multi_Signal_Ballast}
\end{figure}

The integrated multi-signal pruning curve shown in
Fig.~\ref{fig:Multi_Signal_Ballast} combines entropy, mutual
information, topic coherence, redundancy, and information bottleneck
signals to guide optimal threshold selection. Beyond the 0.3-0.4 range,
classification accuracy begins to decline slightly, indicating the
transition from ballast removal to signal loss. Nevertheless, even at
99.4\% feature reduction, model performance drops by only 0.004 from its
peak value, while training time is reduced by nearly 50\%.

These results confirm that ballast in semi-structured datasets is
systemic in nature, encompassing statistical, semantic, and physical
dimensions. The proposed masking framework effectively captures this
heterogeneity, enabling robust and reproducible reductions with minimal
information loss.

This curve demonstrates performance impact of pruning based on entropy
and mutual information thresholds.

A quadrant-based diagnostic for identifying trivial vs. informative
tokens.

Entropy shows variation, MI shows predictive relevance. Their
intersection reveals true signal by filtering out both constant,
low-variation terms and misleading, non-predictive noise.
The entropy-MI plane makes the ballast visible.

Histogram shows co-occurrence redundancy is minimal, supporting
semantic-level filtering.

This curve illustrates the trade-off between feature retention and model
performance when applying increasing thresholds to multi-signal ballast
pruning (TF-IDF, Mutual Information, Topic Coherence, Redundancy, and
Information Bottleneck). As pruning threshold rises, the number of retained features
drops sharply, while accuracy and F1 scores initially remain stable,
then degrade.

\subsection{Ballast Detection in Sparse Data}

The sparse dataset, derived from 2022 Ireland Census, exhibits a
high-dimensional structure (823 columns) with significant sparsity.
After standard cleaning and integration, overall sparsity remained
nearly unchanged (13.69\% $\rightarrow$ 14.17\%), indicating that conventional
pre-processing fails to reduce ballast. Over 50 features exceeded 70\%
sparsity as some nearing 90\%, suggesting low structural utility. These
persisted despite deduplication and formatting corrections, highlighting
structural ballast. A subset of sparse features was isolated for
targeted reduction. No cardinality or variance pruning was applied at
this stage, preserving redundancy and underscoring the need for advanced
ballast-aware techniques.

\emph{1) Profiling Sparse Ballast}

A focused analysis was conducted using both numerical (\textit{T5\_2\_TP})
and categorical (\textit{UR\_Category}) targets. Each sparse feature was
profiled for sparsity, uniqueness, variance, correlation (Pearson,
Spearman), and mutual information (MI). Results showed consistently high
sparsity (\textgreater70\%) and low MI (\textless0.04), with negligible
correlation to targets. Most features exhibited low variance and weak
dependencies as clear indicators of ballast. Visualizations (heatmaps,
scatter plots) confirmed that most features fell into the
high-sparsity/low-signal quadrant. Only a few, such as
\textit{T5\_2\_GE8PH} and \textit{T9\_2\_PI}, showed moderate
informativeness. Some high correlations with \textit{UR\_Category} were
tied to spatial or ID-like fields, raising leakage concerns. These
findings affirm that sparse datasets often contain substantial ballast,
requiring targeted dimensionality reduction.

\emph{2) Ballast Reduction Pipeline}\\
A multi-stage pipeline was applied to a reduced sparse subset
($13{,}185 \times 47$). Initial variance and correlation filters removed
static and redundant features. Mutual information and entropy filters
preserved only features with meaningful signal diversity. LassoCV further
pruned low-impact variables by shrinking coefficients to zero. Finally,
SHAP (Random Forest-based) ranked the remaining features by model
relevance.

As summarized in Table~\ref{tab:feature_reduction_steps}, this process
reduced the original feature set by 91\%, yielding a compact and
informative subset of four features for downstream modelling.

\begin{table}[t]
\caption{Feature reduction steps and remaining feature counts}
\label{tab:feature_reduction_steps}
\centering
\small
\begin{tabular}{p{2.5cm} p{3.6cm} p{1.5cm}}
\hline
\textbf{Step} & \textbf{Description} & \textbf{Features Remaining} \\
\hline
Raw Dataset & Initial feature set & 47 \\
Variance Threshold & Removed near-zero variance features & 46 \\
Correlation Filter & Dropped highly correlated features ($r > 0.95$) & 42 \\
Mutual Information & Retained features informative to the target variable & 18 \\
Entropy Filter & Confirmed relevance; no further reduction observed & 18 \\
LassoCV & Regularized linear feature selection & 14 \\
SHAP (Random Forest) & Retained features above mean SHAP importance & 4 \\
\hline
\end{tabular}
\end{table}

\begin{figure}[t]
    \centering
    \includegraphics[width=\linewidth]{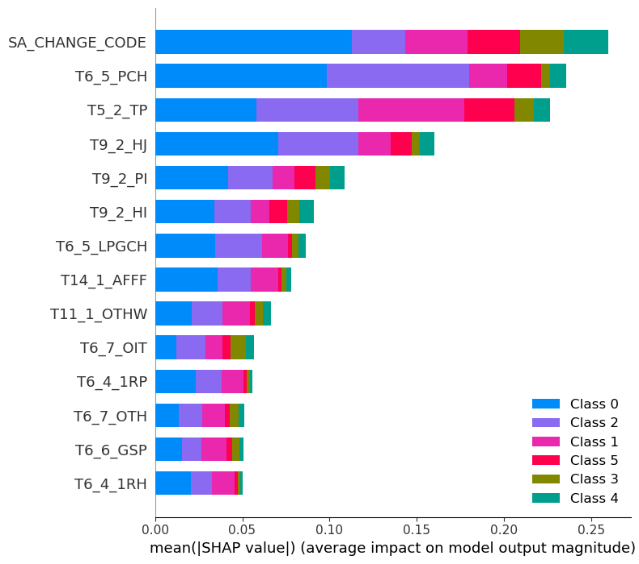}
    \caption{SHAP feature importance bar plot.}
\label{fig:SHAP_plot}
\end{figure}

Fig.~\ref{fig:SHAP_plot} illustrates the average SHAP values of the
14 features retained after Lasso-based selection. Only four features
exceed the mean SHAP importance threshold (highlighted in blue) and are
therefore retained in the final reduced representation.

\emph{3) Ballast Detection \& Pruning (Entropy + MI)}

This phase employed a composite entropy-mutual information framework.
Features were scored using a product-based Ballast Score and pruned
across thresholds from 0\% to 95\%. CatBoost, selected for its robustness
with sparse and categorical data, was used to evaluate classification and
regression performance.

\begin{figure}[t]
    \centering
    \includegraphics[width=\linewidth]{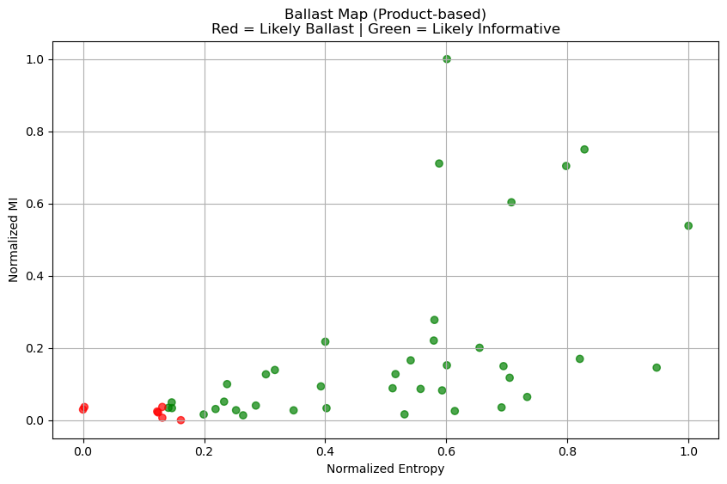}
    \caption{Normalized entropy vs. normalized MI scatterplot, coloured by ballast score.}
\label{fig:entropy_MI_ballastscore}
\end{figure}

Fig.~\ref{fig:entropy_MI_ballastscore} visualizes the joint distribution
of normalized entropy and mutual information, with features coloured by
their ballast scores. A clear separation emerges between informative
features and high-ballast candidates.

\begin{figure}[t]
    \centering
    \includegraphics[width=\linewidth]{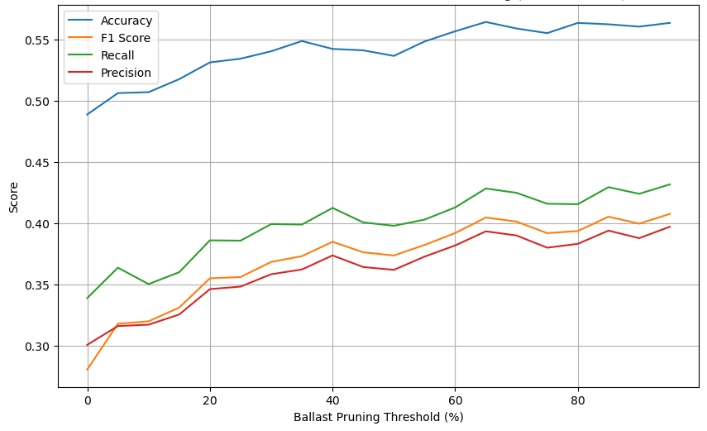}
    \caption{Trade-off curve: performance vs ballast pruning threshold.}
\label{fig:Tradeoff_curve}
\end{figure}

The resulting performance trade-offs are shown in
Fig.~\ref{fig:Tradeoff_curve}, where predictive metrics improve
significantly as pruning increases, peaking at approximately
85-86\% feature removal. Beyond this point, over-pruning leads to a
gradual degradation in generalization performance.

\begin{figure}[t]
    \centering
    \includegraphics[width=\linewidth]{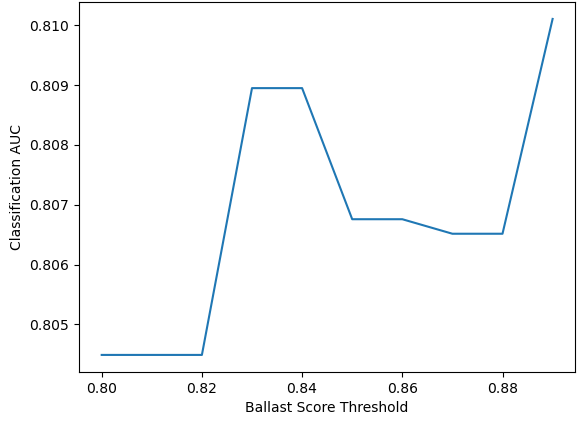}
    \caption{Performance stability across ballast score thresholds between 0.80-0.89.}
\label{fig:ballastscore_thresholds}
\end{figure}

Performance stability across the critical threshold region is examined
in Fig.~\ref{fig:ballastscore_thresholds}. Stable behaviour is observed
between thresholds 0.80 and 0.82, while threshold 0.84 marks a noticeable
increase in AUC and recall. Based on these trends, a final pruning
threshold of 0.86 was selected, retaining 39 high-value features and
maximizing AUC (0.81) while minimizing model complexity.

The observed stability across aggressive pruning thresholds suggests
that most features contribute marginal signal, reinforcing the need for
multi-criteria ballast detection. This enables principled dimensionality
reduction while preserving model robustness and generalization.

\emph{4) Ballast Detection \& Pruning (PCA, VarianceThreshold, Correlation)}

\begin{figure}[t]
    \centering
    \includegraphics[width=\linewidth]{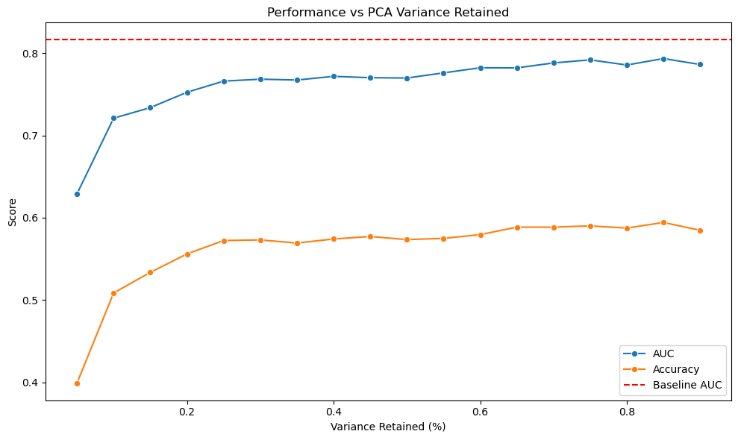}
    \caption{Effect of PCA-based dimensionality reduction on Random Forest
performance.}
\label{fig:PCA_reduction_RandomForest}
\end{figure}

PCA exhibits a clear trade-off between dimensionality and predictive
performance, as shown in Fig.~\ref{fig:PCA_reduction_RandomForest}.
Retaining 85\% of the variance (26 components) preserves a high AUC
($\sim$0.793), confirming that a substantial portion of ballast can be
compressed without measurable performance loss.

\begin{figure}[t]
    \centering
    \includegraphics[width=\linewidth]{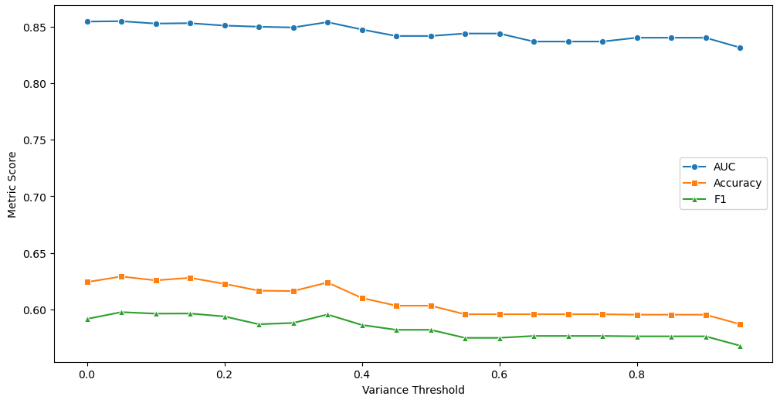}
    \caption{Variance Threshold: Metrics vs. Threshold.}
\label{fig:VarianceThreshold_Metrics_Threshold}
\end{figure}

\begin{figure}[t]
    \centering
    \includegraphics[width=\linewidth]{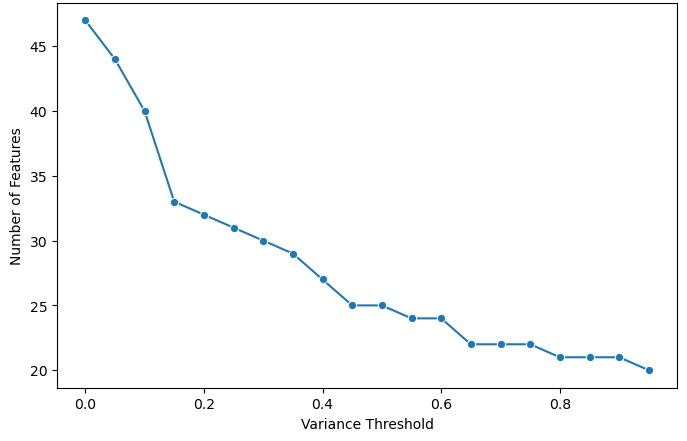}
    \caption{Variance Threshold: Features Remaining vs. Threshold}
\label{fig:VarianceThreshold_FR_Threshold}
\end{figure}

\begin{figure}[t]
    \centering
    \includegraphics[width=\linewidth]{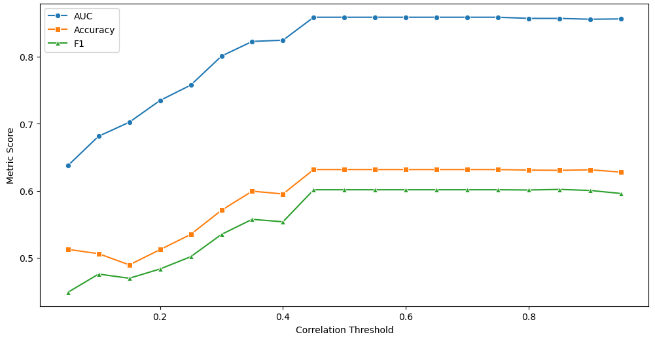}
    \caption{Correlation Filter - Metrics vs. Threshold}
\label{fig:CorrFilter_Metrics_Threshold}
\end{figure}

\begin{figure}[t]
    \centering
    \includegraphics[width=\linewidth]{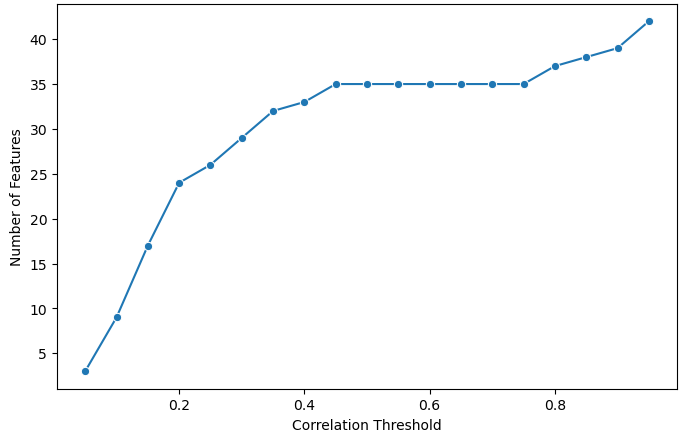}
    \caption{Correlation Filter - Features Remaining vs. Threshold.}
\label{fig:CorrFilter_FR_Threshold}
\end{figure}

VarianceThreshold filtering was effective up to thresholds of
0.10-0.14 (Fig.~\ref{fig:VarianceThreshold_Metrics_Threshold} - \ref{fig:VarianceThreshold_FR_Threshold}), while correlation-based
pruning peaked at thresholds $\sim$0.44-0.46, retaining $\sim$35
features and achieving top AUC ($\sim$0.859,
Fig.~\ref{fig:CorrFilter_Metrics_Threshold} - \ref{fig:CorrFilter_FR_Threshold}). Across methods, the optimal
post-pruning feature count converged to 32-35, indicating a robust
range for ballast-free, performant representations.

Fig.~\ref{fig:PCA_reduction_RandomForest} shows how Random Forest performance metrics (AUC and Accuracy) vary with different levels of PCA variance retention. The red dashed
line represents the baseline AUC using all 47 features (no
dimensionality reduction). As retained variance increases, performance
steadily improves, with a near-optimal AUC reached at 85\% variance
retained (26 components).

Fig.~\ref{fig:VarianceThreshold_Metrics_Threshold} displays how AUC, accuracy, and F1 score change as features with increasing variance thresholds are removed. Performance remains stable
in the 0.05-0.15 range but declines beyond 0.20, suggesting valuable
features begin to be excluded.

Number of features (Fig.~\ref{fig:VarianceThreshold_FR_Threshold}) retained decreases gradually with rising variance
threshold. Optimal trade-off is seen around threshold 0.10-0.14,
retaining 33-44 features.

AUC, accuracy, and F1 (Fig.~\ref{fig:CorrFilter_Metrics_Threshold}) improve with correlation thresholds up to 0.44. Best performance is achieved around 0.42-0.46, indicating effective
removal of redundant features without losing signal.

At correlation thresholds $\geq$~0.40, approximately 35 features
(Fig.~\ref{fig:CorrFilter_FR_Threshold}) remain, corresponding to peak
model performance. Below 0.30, too many features are removed, harming
predictive power.

\emph{5) Ballast Detection \& Pruning (Lasso, SHAP, and Information Bottleneck)}

L1-based pruning revealed a sweet spot at $\alpha \approx$ 0.001,
retaining $\sim$34-41 features with minimal AUC loss ($\sim$0.85).
SHAP pruning with CatBoost showed stable performance up to 50\% feature
removal, peaking at AUC$\approx$0.805 at a threshold of 0.01
(Table~\ref{tab:shap_lgbm_catboost}). Finally, IB pruning identified
ballast via normalized mutual information, yielding optimal
classification (AUC$\approx$0.81) and regression performance when
37-41 informative features were retained. All methods confirm that
judicious ballast reduction can streamline feature space without
compromising predictive power.

\begin{table}[t]
\caption{Comparison of SHAP-based feature selection: \\ LightGBM vs. CatBoost}
\label{tab:shap_lgbm_catboost}
\centering
\small
\begin{tabular}{p{3.0cm} p{2.4cm} p{2.2cm}}
\hline
\textbf{Aspect} & \textbf{LightGBM} & \textbf{CatBoost} \\
\hline
AUC (All Features) & $\sim$0.74 & $\sim$0.80 \\
AUC (After Pruning) & $\sim$0.74 (best at 21-28 features) & 
$\sim$0.80-0.81 (best at $\sim$22 features) \\
F1 Score & $\sim$0.77 & $\sim$0.78 \\
Precision / Recall & $\sim$0.81 / $\sim$0.81 & $\sim$0.82 / $\sim$0.82 \\
SHAP Feature Stability & Moderate fluct. under pruning & Stable up to $\sim$50\% feature removal \\
Feature Importance Spread & Sharp drop-off after top-ranked features &
More inform., smoother top-feature distrib. \\
\hline
\end{tabular}
\end{table}

Table~\ref{tab:shap_lgbm_catboost} presents a comparative analysis of
SHAP-based feature pruning in LightGBM and CatBoost models. The
evaluation includes AUC, F1 score, Precision/Recall, and qualitative
assessments of feature importance distribution and pruning stability.
The results indicate that CatBoost exhibits superior robustness and
interpretability under aggressive feature pruning.

The advantages of the entropy + MI method include balancing sparsity
awareness (entropy) and target informativeness (MI), providing
interpretable pruning via two-dimensional scatter plots and score
thresholds, exhibiting a smooth and stable performance curve across
pruning thresholds (0.80-0.89), and offering easily customizable
thresholds.

\emph{6) Typology of Ballast Information in Sparse Structured}

Classification of ballast types in sparse datasets along with their characteristic indicators and corresponding detection or pruning methods, is presented in Table~\ref{tab:ballast_typology_sparse}. Each ballast category aligns with a specific set of statistical, structural, or model-aware techniques suitable for effective reduction.

\begin{table*}[t]
\caption{Typology of ballast information in sparse dataset}
\label{tab:ballast_typology_sparse}
\centering
\small
\begin{tabular}{p{3.5cm} p{4.8cm} p{4.1cm} p{4.0cm}}
\hline
\textbf{Ballast Type} &
\textbf{Description} &
\textbf{Indicators} &
\textbf{Detection / Removal Methods} \\
\hline
High-sparsity features & Features with extreme missing or zero values &
Sparsity $>$70--90\%; many nulls or default encodings & Sparsity Filter, Entropy Filter \\

Low-variance features & Features with nearly constant values across records &
Variance near zero; no discriminative power & VarianceThreshold \\

Redundant features & Features highly correlated with others (linear or monotonic) &
Pearson/Spearman $\rho > 0.95$ & Correlation Filter \\

Low mutual information & Features weakly associated with target labels &
MI $<$ 0.03-0.05 & Mutual Information (MI), Entropy--MI Scoring \\

Structural artifacts & System-generated identifiers or placeholders &
Uniform patterns, IDs, spatial zones without predictive value &
SHAP, Domain Filtering, Data Leakage Checks \\

Over-encoded categories & Sparse one-hot or multi-hot encodings of categorical fields &
High dimensionality with sparse representation & PCA, Embedding, SHAP, Lasso \\

Contextually irrelevant data & Features unrelated to the modelling task (e.g., metadata, survey structure) & No logical link to target; fails feature importance tests &
LassoCV, SHAP, Information Bottleneck (IB) \\
\hline
\end{tabular}
\end{table*}

\subsection{Mathematical Model of Ballast Information in\\ 
Multimodal Datasets}

\emph{1) Ballast definition and core hypotheses}

Let $D = \{X, y\}$ be a multimodal dataset with $n$ samples and $m$ features or data elements,
where $X \in \mathbb{R}^{n \times m}$ and $y \in \mathcal{Y}$ is a task-dependent target
(classification, regression, or clustering).
In this framework, \textit{ballast} refers to features or tokens that are structurally present
but contribute\textit{ minimal analytical value}. Based on the multimodal study, four primary ballast types are distinguished:
\begin{itemize}
    \item \textit{Statistical Ballast}: Elements with low variance or entropy,
    indicating uniformity or noise.
    
    \item \textit{Redundancy Ballast}: Elements strongly correlated or semantically
    overlapping with others, offering duplicated information.
    
    \item \textit{Semantic Ballast}: In unstructured or semi-structured data,
    elements with low thematic coherence or low embedding information density.
    
    \item \textit{Model-Irrelevant Ballast}: Features with negligible influence
    on model predictions (e.g., low SHAP value, Lasso zero-coefficient).
\end{itemize}

\textit{Hypothesis:}
Ballast arises from combinations of low utility (informativeness) and high redundancy
and varies by modality and task. A general mathematical framework must normalize and integrate these signals to support robust, cross-modal ballast detection.

\emph{2) Ballast Index (Feature-Level)}

The \textit{BallastScore} is defined for each feature or token $X_j$ as a weighted sum
of its utility deficits and redundancy indicators:
\begin{equation}
\mathrm{\textit{BallastScore}}_j =
\sum_{k=1}^{K} w_k \cdot (1 - U_{jk})
+
\sum_{r=1}^{R} \lambda_r \, R_{jr},
\label{eq:ballast_score}
\end{equation}

where:
\begin{itemize}
    \item $U_{jk} \in [0,1]$: normalized utility signals (lower values imply lower informativeness),
    \item $R_{jr} \in [0,1]$: normalized redundancy metrics (higher values imply stronger redundancy),
    \item $w_k, \lambda_r$: scalar weights reflecting the importance of each signal,
    \item $K$: number of utility metrics (e.g., mutual information, entropy, variance, topic coherence, SHAP),
    \item $R$: number of redundancy metrics (e.g., correlation, cosine similarity, embedding overlap).
\end{itemize}

\textit{Interpretation:}
\begin{itemize}
    \item A high \textit{BallastScore} (closer to 1) indicates low utility and/or high redundancy
    and thus represents a strong ballast candidate.
    \item A low \textit{BallastScore} (closer to 0) implies that the feature is both informative
    and non-redundant.
\end{itemize}

This formulation supports features from structured (tabular), semi-structured
(e.g., JSON, XML), unstructured (text or image), and sparse datasets by normalizing
all signals to the range $[0,1]$.

\emph{3) Dataset-Level Ballast Index}

The \textit{dataset-level ballast index} $B(D)$ is defined as the fraction of features
classified as ballast based on a threshold $\tau \in [0,1]$:
\begin{equation}
B(D) = \frac{1}{m} \sum_{j=1}^{m} \mathbb{I}\left[\text{\textit{BallastScore}}_j > \tau \right],
\label{eq:ballast_index}
\end{equation}
where $\mathbb{I}[. . .]$ is the indicator function, and
$\tau$ is an empirically calibrated threshold (e.g., based on inflection points
in model performance, AUC curves, or SHAP drop-offs).

This index quantifies the \textit{proportion of ballast within a dataset} and can be
reported per modality or feature group to localize ballast hotspots.

\emph{4) Pruning Criterion}

The ballast pruning operator $\Pi_{\tau}$ is defined as:
\begin{equation}
\Pi_{\tau}(X) = \left\{ X_j \in X \mid \text{\textit{BallastScore}}_j \leq \tau \right\},
\label{eq:pruning_operator}
\end{equation}
that is, only features with a \textit{BallastScore} below the threshold $\tau$
are retained. The value of $\tau$ is tuned based on task-specific model performance.

This formal model quantifies ballast through a modular, interpretable index
integrating four core dimensions: \textit{statistical insignificance, redundancy,
semantic incoherence,} and model \textit{irrelevance}. By enabling multi-signal detection
and cross-modal generalization, it provides a principled basis for task-aware
data reduction in large-scale, heterogeneous datasets.

\section{CONCLUSIONS AND FUTURE WORK}

This study presents a comprehensive, multimodal framework for ballast
detection and reduction across structured, semi-structured,
unstructured, and sparse datasets. By integrating statistical, semantic,
and model-aware signals, a generalized \emph{\textbf{Ballast Score}} is
established, and its effectiveness is demonstrated across diverse data
regimes. The key finding is that ballast corresponds to structurally
present yet analytically low-utility data, which can be systematically
quantified and pruned with minimal degradation to downstream model
performance. In multiple scenarios, including SHAP-based pruning for
structured data and entropy-mutual information (MI) filtering for sparse
and unstructured data, performance gains and significant reductions in
computational or storage overhead are achieved. Collectively, these
findings confirm that ballast represents a quantifiable and removable
form of data inefficiency rather than random noise.

A substantial share of ballast in datasets is believed to originate from
unconscious human decisions during system design and data collection.
When creating schemas or logging systems, humans rarely foresee all
future analytical needs, leading to the inclusion of fields that are
redundant, sparsely populated, or irrelevant. This cognitive gap as
driven by overengineering, organizational habits, or short-term tasks,
results in semantically and statistically low-value data. Over time,
such features persist without re-evaluation, especially in complex
systems with multiple stakeholders.

Practically, this research has direct implications for large-scale data
processing, particularly in domains where high-dimensionality or
semantic redundancy impairs model interpretability, computational
efficiency, and storage scalability. Additionally, modality-specific
ballast typology (e.g., statistical ballast in structured data, semantic
redundancy in unstructured text, or infrastructural ballast in sparse
formats) provides actionable insights in selecting appropriate pruning
strategies.

However, the study is not without limitations. While the proposed
ballast index generalizes across data modalities, its calibration (e.g.,
threshold $\tau$) remains dataset- and task-dependent. Moreover, SHAP
and Lasso underperform in semantically complex or lexically noisy
settings, such as OCR text and scholarly corpora, where signal is not
linearly separable. Furthermore, although entropy and MI perform
robustly in unstructured contexts, their class-agnostic nature may omit
low-frequency but high-value features in imbalanced tasks.

Future work should focus on two directions. First, extending ballast
framework to non-textual modalities such as images, graphs, and audio,
where redundancy and informativeness manifest differently, would
validate its generality. Second, dynamic or adaptive thresholding
strategies for the Ballast Score, perhaps informed by performance
plateaus, SHAP-gradient inflection or by some other method, could yield
more robust and context-aware pruning.

\emph{Brief summary of the study:}

1. Different methods reveal different facets of ballast.

2. Ballast is task-dependent.

3. Entropy and MI are reliable starting points. The entropy-MI plane makes the ballast visible.

4. Semantic pruning adds interpretive depth, but not always practical.

5. Reducing ballast improves more than just performance, it improves the entire data science workflow.

6. Evaluation needs multi-faceted metrics, including metrics like feature count reduction, training time, and post-pruning interpretability to give a fuller view.

\bibliographystyle{IEEEtran}
\bibliography{sources}

\appendices

\section{Unstructured Dataset (CORD-19): Detailed Methodological Description}
\label{app:detailed_method_descr}

\subsection{Dataset Description and Justification}

The unstructured data analysis was conducted using the CORD-19 dataset, an open-access corpus of scholarly literature related to COVID-19 compiled by the Allen Institute for AI. The dataset contains over one million scientific articles, including approximately 370,000 documents with full-text availability in JSON format. Available metadata includes titles, abstracts, publication information, and structured full-body sections.

The CORD-19 dataset was selected for the following reasons:

\subsubsection*{Semantic complexity}
As a full-text corpus of academic publications, the dataset exhibits diverse linguistic structures, domain-specific terminology, and substantial variation in document length, representing core challenges of unstructured textual data.

\subsubsection*{Volume and variability}
With hundreds of thousands of heterogeneous documents ranging from abstracts to full-length articles exceeding 1,000 sentences, the dataset is representative of real-world large-scale natural language corpora.

\subsubsection*{Scientific relevance}
CORD-19 has been widely adopted in biomedical text mining and information retrieval research, providing a well-established benchmark for methodological evaluation.

\subsubsection*{Data modality}
Unlike structured tabular data, information is encoded in free-form prose, making the dataset particularly suitable for evaluating semantic ballast detection methods based on redundancy, topical irrelevance, and entropy degradation.

\subsection{Data Processing and Analytical Strategy}

The ballast detection pipeline for unstructured data combined shallow statistical heuristics with deep semantic modelling techniques to identify low-information, redundant, or semantically uninformative textual units (ballast sentences) that reduce dataset utility without contributing semantic value.

The pipeline comprised the following stages:

\subsubsection{Data Extraction and Sampling}

Full-text JSON files were parsed, and the following fields were extracted: \textit{title}, \textit{abstract}, and \textit{body\_text}. Articles containing fewer than 100 words in the full-text body were excluded. For computational tractability, a stratified sample of approximately 10{,}000 documents was selected for analysis.

This sampling strategy ensured coverage of diverse document structures while maintaining manageable computational requirements. Short documents were excluded to avoid unstable or misleading entropy and redundancy estimates.

\subsubsection{Text Pre-processing}

Standard natural language processing pre-processing steps were applied:

\begin{itemize}
    \item Lowercasing and stop-word removal using \texttt{NLTK},
    \item Tokenization and sentence segmentation via \texttt{word\_tokenize} and \texttt{sent\_tokenize},
    \item Lemmatization using \texttt{spaCy},
    \item Removal of numeric tokens and punctuation-only tokens.
\end{itemize}

These steps reduced lexical noise and normalized textual representations to enable consistent vectorization and semantic modelling across documents.

\subsubsection{Statistical Ballast Detection (Shallow Methods)}

An initial round of ballast identification employed statistical indicators of information sparsity.

a) \textit{Term Frequency--Inverse Document Frequency (TF-IDF)}

TF-IDF scores were computed across the corpus using \texttt{TfidfVectorizer}. Sentences with aggregate TF-IDF values within the lowest decile of the distribution were flagged as low-information candidates.

b) \textit{Shannon Entropy}

Sentence-level Shannon entropy was calculated based on empirical token frequency distributions:

\begin{equation}
H(S) = -\sum_{i=1}^{n} p(w_i)\log_2 p(w_i)
\end{equation}

where $p(w_i)$ denotes the empirical probability of token $w_i$ within sentence $S$. Sentences with entropy values below 1.5 were considered lexically sparse and marked as ballast candidates.

TF-IDF captures content distinctiveness, while entropy quantifies lexical diversity; both metrics are widely used indicators of information richness in text corpora \cite{Shannon1948,Manning2009}.

\subsection{Semantic Ballast Detection (Deep Methods)}

To refine the ballast filtering process, two semantically grounded techniques were employed.

\subsubsection{Sentence Embeddings and Cosine Similarity}

Sentence embeddings were generated using a Sentence-BERT model (DistilBERT), producing 768-dimensional vector representations. Pairwise cosine similarity was computed both between consecutive sentences and across entire documents. Sentences exhibiting cosine similarity values greater than or equal to 0.95 were flagged as highly redundant.

This step targeted semantic repetition, where sentences contribute minimal novel information despite lexical variation.

\subsubsection{Topic Coherence via Latent Dirichlet Allocation}

Latent Dirichlet Allocation (LDA) models were trained using Gensim with the number of topics set to 20. Each sentence was assigned a topic distribution, and Jensen--Shannon divergence was computed between sentence-level topic vectors and the dominant topic distribution of the corresponding document.

Sentences with high divergence values, indicating low topical coherence, were treated as off-topic ballast \cite{Blei2003,SciPyJS2024}.

\subsection{Aggregation and Filtering Strategy}

Outputs from the four ballast indicators (low TF-IDF, low entropy, high semantic redundancy, and low topic coherence) were aggregated using a voting-based decision rule. A sentence was classified as ballast if at least two of the four criteria were satisfied.

Filtered documents were reconstructed using only retained sentences, producing ballast-reduced textual representations for downstream evaluation.

\subsection{Evaluation of Ballast Removal}

To assess the effectiveness of ballast removal, a LightGBM classifier was trained to categorize documents into biomedical subdomains (e.g., virology, epidemiology) using both original and ballast-reduced text representations.

Evaluation metrics included Accuracy, F1-score, Recall, and ROC-AUC. Across experiments, ballast-reduced representations yielded modest but consistent improvements in classification performance, attributable to reduced semantic noise and redundancy.

Additional descriptive metrics included:
\begin{itemize}
    \item Sentence reduction ratio (approximately 41\% of sentences identified as ballast on average),
    \item Shifts in entropy and topic coherence distributions following filtering.
\end{itemize}

\subsection{Implementation Details}

The following libraries and tools were used:

\begin{itemize}
    \item \texttt{pandas}, \textbf{numpy}: data manipulation and numerical processing,
    \item \texttt{NLTK}, \texttt{spaCy}: text pre-processing,
    \item \texttt{scikit-learn}: TF-IDF computation and evaluation,
    \item \texttt{sentence-transformers}: semantic embeddings,
    \item \texttt{Gensim}: LDA topic modelling,
    \item \texttt{LightGBM}, \texttt{SHAP}: downstream classification and interpretability,
    \item \texttt{matplotlib}, \texttt{seaborn}: visualization.
\end{itemize}

\subsection{Summary}

This\textit{ Appendix~A} demonstrates that the proposed ballast detection framework can be systematically instantiated for large-scale unstructured text corpora by integrating statistical sparsity measures with semantic redundancy and topical coherence analysis. The presented configuration serves as a reference implementation for unstructured data modalities and complements the high-level methodology described in the main body of the paper.

\section{Ballast Taxonomy Table (\ref{tab:ballast_methods_comparison})}

\begin{table*}[t]
\centering
\caption{Comparative effectiveness of ballast detection methods across data modalities}
\label{tab:ballast_methods_comparison}
\footnotesize
\setlength{\tabcolsep}{4pt}
\renewcommand{\arraystretch}{1.0}
\begin{tabular}{p{2.8cm} p{3.4cm} p{4.0cm} p{3.8cm} p{3.4cm}}
\hline
\textbf{Method} &
\textbf{Structured Dataset} &
\textbf{Unstructured Dataset} &
\textbf{Semi-Structured Dataset} &
\textbf{Sparse Dataset} \\
\hline

\textbf{\textit{Mutual Information (MI)}} &
Detected statistically irrelevant features with weak target signal (e.g., price, category); rejected approximately 80 features. &
Effective in isolating boilerplate and citation-like structures in CORD-19; also flagged low-content layout regions in PubLayNet. &
Highlighted weakly informative textual fields and categorical encodings; effective for numerical metadata. &
Rejected fields with MI $< 0.03$; highly effective for dimensionality reduction while preserving predictive signal. \\

\textbf{\textit{Entropy}} &
Detected flat distributions; flagged features with uniform value spread (e.g., device information). &
Exposed repeated tokens and layout blocks in PubLayNet; revealed document templates in both datasets. &
Identified repeated tokens in descriptions and metadata; reinforced MI-based filtering. &
Detected zero-variance and ultra-sparse columns; strong synergy with MI for structure-first filtering. \\

\textbf{\textit{SHAP}} &
Ranked features by model relevance (LightGBM); four features retained post-processing; best interpretability--performance trade-off. &
Performed poorly on text-dominant data due to limited structural grounding; weak thematic alignment. &
Weak correlation with semantic methods; struggled with high-dimensional sparse textual inputs. &
Final reduction stage; retained top four predictors based on SHAP values; excellent interpretability. \\

\textbf{\textit{Lasso}} &
Efficient linear pruning; eliminated zero-contributing fields (105 $\rightarrow$ 14 features). &
Missed nonlinear ballast in free-text; underperformed compared to semantic detectors such as LDA and BERT. &
Detected structurally redundant fields; insensitive to lexical overlap; effective for numerical variables. &
Effective after MI and entropy filtering; removed redundant encodings and noise features. \\

\textbf{\textit{PCA / t-SNE}} &
Filtered latent ballast; achieved best AUC (0.998) but sacrificed interpretability (432 $\rightarrow$ 23 features). &
Limited explainability for text; captured both noise and signal; effective only for embedding-based representations. &
Detected over-encoded sparse fields; useful for visualizing low-dimensional semantic clusters. &
Detected latent sparsity but removed interpretable signals; unsuitable for end-to-end model explanation. \\

\textbf{\textit{Correlation Filter}} &
Mitigated multicollinearity (432 $\rightarrow$ 116 features) with minimal performance loss. &
Not applicable to raw high-dimensional text without embeddings. &
Detected duplicated structured metadata fields; ineffective for free-text descriptions. &
Removed redundant population statistics using Spearman $\rho > 0.95$; foundational pipeline step. \\

\textbf{\textit{TF-IDF}} &
Not applicable. &
Removed overused generic terms; effective for early-stage ballast detection in CORD-19 and PubLayNet. &
Pruned non-informative tokens in textual descriptions; filtered spam-like reviews. &
Not applied. \\

\textbf{\textit{LDA (Topic Coherence)}} &
Not applicable. &
Detected topic-incoherent segments; optimal threshold at 0.30 for CORD-19; stable performance on PubLayNet. &
Identified noisy review blocks lacking topic affiliation; effective for semantic ballast classification. &
Not applied. \\

\textbf{\textit{BERT Similarity}} &
Not applicable. &
Detected semantically repetitive content in CORD-19; later activation in PubLayNet (threshold $\geq 0.85$). &
Identified redundant or copy-pasted user reviews; correlated with embedding-based clustering. &
Not applied. \\

\textbf{\textit{Clustering (K-Means, DBSCAN)}} &
Not applicable. &
Identified text segments forming noisy, topic-independent clusters in both datasets. &
Separated semantically weak clusters in review text; supported embedding-based filtering. &
Not applied. \\

\textbf{\textit{Information Bottleneck}} &
Reduced features to 80; moderate AUC (0.72); effective as a compression-aware auxiliary method. &
Captured low-contribution terms; non-monotonic gains; effective for layout-based ballast in PubLayNet. &
Assisted MI and entropy methods; aided calibration of pruning thresholds. &
Flagged features with residual contribution below $\tau = 0.30$; supported by MI and entropy rankings. \\

\textbf{\textit{IoU-Based Redundancy}} &
Not applicable. &
Highly effective for PubLayNet (IoU $> 0.64$ at $\tau = 0.85$); flagged redundant layout components. &
Weak signal; primarily useful for repeated field positions. &
Not applied. \\

\hline
\end{tabular}
\end{table*}

Table~\ref{tab:ballast_methods_comparison} summarizes the comparative performance of ballast detection methods across structured, unstructured, semi-structured, and sparse datasets.

\textbf{Interpretation}

\begin{itemize}
\item \emph{Structured Dataset:} Lasso, SHAP, and Correlation Filters were
the most effective. PCA showed peak performance but with reduced
interpretability. SHAP best balanced feature importance and
interpretability. MI and Entropy provided early detection of
non-contributory features.

\item \emph{Unstructured Dataset:} Entropy and MI outperformed others by
identifying statistically and structurally redundant text. BERT and LDA
provided complementary semantic ballast detection, while TextRank and
IoU handled layout-based or templated ballast effectively.

\item \emph{Semi-Structured Dataset:} The best outcomes emerged from hybrid
combinations as TF-IDF and LDA for semantic ballast, MI and Entropy for
structural ballast, and SHAP for model-awareness. PCA was used
cautiously to compress over-encoded metadata.

\item \emph{Sparse Dataset:} A multi-stage pipeline proved necessary.
Entropy and MI led the reduction phase, followed by Lasso and SHAP for
final selection. PCA and t-SNE were informative but not optimal for
production modelling due to interpretability loss.

\end{itemize}

\section{Condensed Ballast Typology Table (\ref{tab:condensed_ballast_typology})}

For clarity, a condensed ballast typology is summarized in Table~\ref{tab:condensed_ballast_typology}.

\begin{table*}[t]
\centering
\caption{Condensed ballast typology and corresponding detection methods}
\label{tab:condensed_ballast_typology}
\footnotesize
\setlength{\tabcolsep}{6pt}
\renewcommand{\arraystretch}{1.15}
\begin{tabular}{p{2.9cm} p{5.4cm} p{5.8cm} p{3.0cm}}
\hline
\textbf{Ballast Type} &
\textbf{Detection Methods} &
\textbf{Description} &
\textbf{Effective In} \\
\hline

\textbf{\textit{Statistical Ballast}} &
Entropy, Mutual Information (MI), Variance Threshold &
Low-variability or uniform features with minimal discriminative power. &
Structured, Sparse, Semi-Structured \\

\textbf{\textit{Redundancy Ballast}} &
Correlation Filter, Lasso, SHAP, PCA, IoU &
Features duplicating others either numerically or structurally, such as collinearity or templated layouts. &
Structured, Sparse, Unstructured \\

\textbf{\textit{Semantic Ballast}} &
TF-IDF, LDA (Topic Modelling), BERT Similarity, Clustering (K-Means, DBSCAN) &
Lexically or thematically non-informative components, including boilerplate text, common phrases, and repetitive segments. &
Unstructured, Semi-Structured \\

\textbf{\textit{Model-Irrelevant Ballast}} &
SHAP, Lasso, Information Bottleneck &
Features exhibiting negligible contribution to model outputs, such as near-zero coefficients or low SHAP values. &
Structured, Sparse, Semi-Structured \\

\hline
\end{tabular}
\end{table*}

\end{document}